\documentclass[journal]{IEEEtran}
\usepackage{multirow}
\usepackage{subcaption}
\usepackage{booktabs}
\ifCLASSINFOpdf
   \usepackage[pdftex]{graphicx}
\else
\fi
\usepackage{amsmath}
\usepackage{amsfonts}
\usepackage[ruled,linesnumbered]{algorithm2e}
\usepackage{algorithmic}
\begin{document}
%
\title{STJLA: A Multi-Context Aware \underline{S}patio-\underline{T}emporal \underline{J}oint \underline{L}inear \underline{A}ttention Network for Traffic Forecasting}
%
%
%
\author{Yuchen~Fang,
        Yanjun~Qin,
        Haiyong~Luo,
        Fang~Zhao,
        and~Chenxing~Wang
\thanks{Manuscript received Jun 18, 2021. This work was supported in part by the National Key Research and Development Program under Grant 2018YFB0505200, the Action Plan Project of the Beijing University of Posts and Telecommunications supported by the Fundamental Research Funds for the Central Universities under Grant 2019XD-A06, the Special Project for Youth Research and Innovation, Beijing University of Posts and Telecommunications, the Fundamental Research Funds for the Central Universities under Grant 2019PTB-011, the National Natural Science Foundation of China under Grant 61872046, the Joint Research Fund for Beijing Natural Science Foundation and Haidian Original Innovation under Grant L192004, the Key Research and Development Project from Hebei Province under Grant 19210404D and 20313701D, the Science and Technology Plan Project of Inner Mongolia Autonomous Region under Grant 2019GG328 and the Open Project of the Beijing Key Laboratory of Mobile Computing and Pervasive Device. (Corresponding author: Haiyong Luo; Fang Zhao.)}
\thanks{Yuchen Fang, Yanjun Qin, Fang Zhao, and Chenxing Wang are with the School of Computer Science (National Pilot Software Engineering School), Beijing University of Posts and Telecommunications, Beijing 100876, China (e-mail: fangyuchen@bupt.edu.cn; qinyanjun@bupt.edu.cn; zfsse@bupt.edu.cn; wangchenxing@bupt.edu.cn).}
\thanks{Haiyong Luo is with the Research Center for Ubiquitous Computing Systems, Institute of Computing Technology Chinese Academy of Sciences, Beijing 100190, China (e-mail: yhluo@ict.ac.cn).}}

%
%

\markboth{IEEE TRANSACTIONS ON INTELLIGENT TRANSPORTATION SYSTEMS}%
{Fang \MakeLowercase{\textit{et al.}}: STJLA: A Multi-Context Aware \underline{S}patio-\underline{T}emporal \underline{J}oint \underline{L}inear \underline{A}ttention Network for Traffic Forecasting}
%



\maketitle

\begin{abstract}
Traffic prediction has gradually attracted the attention of researchers because of the increase in traffic big data. Therefore, how to mine the complex spatio-temporal correlations in traffic data to predict traffic conditions more accurately become a difficult problem. Previous works combined graph convolution networks (GCNs) and self-attention mechanism with deep time series models (e.g. recurrent neural networks) to capture the spatio-temporal correlations separately, ignoring the relationships across time and space. Besides, GCNs are limited by over-smoothing issue and self-attention is limited by quadratic problem, result in GCNs lack global representation capabilities, and self-attention inefficiently capture the global spatial dependence. In this paper, we propose a novel deep learning model for traffic forecasting, named Multi-Context Aware \underline{S}patio-\underline{T}emporal \underline{J}oint \underline{L}inear \underline{A}ttention (STJLA), which applies linear attention to the spatio-temporal joint graph to capture global dependence between all spatio-temporal nodes efficiently. More specifically, STJLA utilizes static structural context and dynamic semantic context to improve model performance. The static structure context based on node2vec and one-hot encoding enriches the spatio-temporal position information. Furthermore, the multi-head diffusion convolution network based dynamic spatial context enhances the local spatial perception ability, and the GRU based dynamic temporal context stabilizes sequence position information of the linear attention, respectively. Experiments on two real-world traffic datasets, England and PEMSD7, demonstrate that our STJLA can achieve up to 9.83\% and 3.08\% accuracy improvement in MAE measure over state-of-the-art baselines.
\end{abstract}


%
\IEEEpeerreviewmaketitle

\section{Introduction}
%
%
%
%
\IEEEPARstart{T}{he} gradually increasing traffic demand and population has brought challenges to traffic management. Intelligent transportation systems (ITS) \cite{dimitrakopoulos2010intelligent} have attracted researchers' attention and the traffic forecasting is a very important and challenging task in intelligent transportation systems. The goal of traffic forecasting is to predict a period of future traffic data through a period of known historical traffic data. Accurate and efficient prediction of future traffic data can assist traffic management departments better schedule traffic to avoid congestion. 
\begin{figure}
    \centering
    \includegraphics[width=1.0\linewidth]{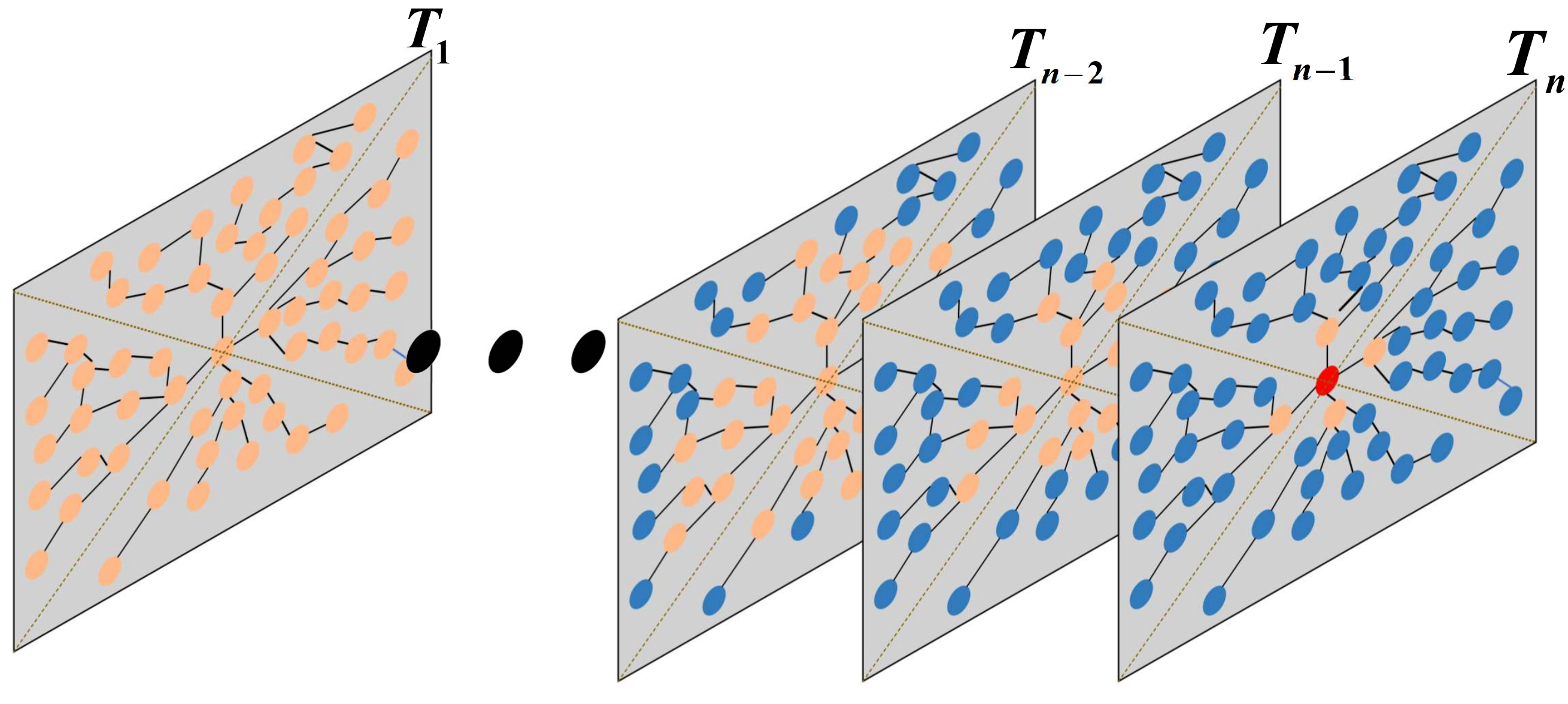}
    \caption{The receptive field (orange nodes) of the red node increases as the time moves backward in the spatio-temporal graph.}
\end{figure}

It is well known that the traffic forecasting task is very challenging, mainly due to the complex spatio-temporal dependencies. We first review how the previous works dealt with complex spatio-temporal dependencies. The early traditional statistical methods ARIMA \cite{williams2003modeling}, shallow machine learning methods SVR \cite{wu2004travel}, and deep time series models LSTM \cite{zhao2017lstm} just set out to extract linear and non-linear temporal dependence and ignored spatial correlations. DCRNN \cite{DBLP:conf/iclr/LiYS018} and STGCN \cite{DBLP:conf/ijcai/YuYZ18} noticed the shortcomings of the above methods and introduced graph convolution networks (GCNs) \cite{bruna2013spectral} into traffic forecasting to extract the spatial dependence. They combine GCNs and deep time series models (e.g. recurrent neural networks (RNNs) \cite{DBLP:conf/adma/WuHSCL17} and convolution neural networks (CNNs) \cite{o2015introduction}) to extract spatio-temporal dependencies separately. We call this spatio-temporal mining mode GCN+TIME. GCN determines how much information the node should obtain from other nodes at the current time, and TIME determines how much the node should obtain from its past. The models GWN \cite{DBLP:conf/ijcai/WuPLJZ19}, MTGNN \cite{wu2020connecting} and AGCRN \cite{DBLP:conf/nips/0001YL0020} based on these two works have not departed from the way of extracting spatio-temporal dependencies using GCN+TIME. Later models such as GMAN \cite{zheng2020gman} used a self-attention (SA) mechanism \cite{DBLP:conf/nips/VaswaniSPUJGKP17} instead of GCN in order to dynamically extract the global dependence, which we call SA+TIME.

We think about why SA+TIME can achieve better results than GCN+TIME, is it global, dynamic, or both? The relationship between nodes is always dynamic. There is not much discussion about global. Why do we need global spatial dependence? As Fig. 1 shows, we found that the historical spatial receptive field of nodes becomes larger as time goes back. Because the distance of the vehicle moving in the road network is proportional to the moving time. So the longer the retreat, the larger receptive field is needed to transmit the historical spatial information to the future. SA+TIME achieved better results than GCN+TIME mode because it has a global receptive field at each time slice. However, if the receptive field is too large, the node information that is not needed will be gathered. It is very difficult to select a suitable receptive field under frameworks of GCN+TIME and SA+TIME. Besides, future state have different requirements for historical spatial information, and above frameworks always use GCN or SA to aggregate information first and then stuff the aggregated information into all future states, which difficult to achieve excellent results. Is it possible not to follow the GCN+Time or SA+Time mode? The answer is yes.

In this paper, we propose a novel Multi-Context Aware \underline{S}patio-\underline{T}emporal \underline{J}oint \underline{L}inear \underline{A}ttention Network (STJLA) for traffic forecasting, which design a spatio-temporal joint mode that combines the sub-graphs of each time slice as a graph, which contains all spatio-temporal states. Then we consider utilizing the self-attention mechanism based on the spatio-temporal joint graph to dynamically calculate correlations between all nodes, so that future states has its own receptive field instead of obtaining same historical information. However, the huge number of states in the spatio-temporal joint graph is impossible to use the self-attention mechanism with a complexity of $O(T^2N^2)$. Aiming at the quadratic problem of self-attention, we use linear attention to speed up the calculation and reduce memory consumption, so that we can efficiently calculate the correlations between all nodes. Besides, we extract the static structure context and dynamic semantic context information to increase the performance of our model. The static structure context is composed of the position information in the spatial topology and the stage of the time slice. However, the static spatial structure context is biased because of the prior road network. For example, two sensors at the same intersection may record the flow of different turns. Besides, we only use the static temporal structure context will lose the order of the long sequence. Therefore, we use multi-head diffusion convolution network (MHDCN) based dynamic spatial semantic context and gated recurrent units (GRU) \cite{chung2014empirical} based dynamic temporal semantic context to dynamically capture local spatio-temporal dependencies to make up for the prior biases and augment order information of long sequence. The main contributions of our work are summarized as follows:
\begin{itemize}
\item We propose STJLA, which introduces the linear attention mechanism to model the global correlations explicitly in spatio-temporal joint graph.
\item We leverage static context based on the spatial topological structure and the temporal position structure to enrich the positional information of the linear attention.
\item We use dynamic context based on MHDCN and GRU to enhance the local perception ability of the linear attention.
\item We conducted extensive experiments on two large-scale real-world traffic datasets, i.e., England and PEMSD7, and the proposed approach obtains 9.83\% and 3.08\% improvement over state-of-the-art baseline methods.
\end{itemize}
\section{Related Works}
\subsubsection{Traffic Forecasting}
Traffic forecasting has been studied for decades. At first, people used auto-regressive integrated moving average (ARIMA) \cite{williams2003modeling}, a traditional statistical method based on the assumption of time series stability, to predict traffic. This method only captures linear dependence and cannot achieve excellent results on unstable traffic data. Compared with traditional methods, shallow machine learning methods capture non-linear correlations in data. Wu et al. \cite{wu2004travel} used support vector regression (SVR) and Van et al. \cite{van2012short} used k-nearest neighbors (KNN) to predict traffic. However, shallow machine learning methods need to manually extract high-order features in the data for specific scenarios and lack of generalization. 

In recent years, with the brilliant success of deep learning in computer vision \cite{forsyth2012computer}, natural language processing \cite{liddy2001natural} and other fields \cite{owens1993signal,hand2014data}. Some attempts have been made for traffic forecasting task, they apply RNNs and its variants (e.g. long short-term memory (LSTM) \cite{hochreiter1997long} and GRU) and improve performance compared with above traditional methods. Compared with shallow models, deep neural networks can model more complex time series, but these models still do not consider correlations between sensors in the traffic network. 

One reason for the outstanding success of CNNs in computer vision is because of its spatial aggregation ability, i.e., it can aggregate the information from other pixels in the receptive field to capture the spatial dependence. Because of this, for the first time, Zhang et al. \cite{zhang2016dnn} used CNNs for image-based traffic prediction tasks to extract spatial correlations in pixels. Subsequently, ConvLSTM \cite{liu2017short} combines convolution with LSTM to capture longer range temporal dependence, and ST-ResNet \cite{zhang2017deep} uses residual neural networks to increase the depth of the model to capture the global spatial dependence. Although CNNs have achieved excellent performance in image-based traffic prediction tasks, it is not suitable for road network-based traffic prediction tasks because the road network is not Euclidean.

Graph convolution operation is a convolution operation defined for non-Euclidean space \cite{zhou2020graph}. It is consistent with the role of CNNs on the image capture spatial correlations; it captures the spatial dependence in the non-Euclidean structure by passing messages between nodes. DCRNN applied GCN to road network-based traffic prediction tasks for the first time, and they have improved its performance compared to previous works. DCRNN replaces the linear operation in the RNN with the graph diffusion convolution \cite{atwood2016diffusion}, which recurrently uses the graph convolution to extract the spatial dependence in each time slice. Another strategy to use graph convolution is to combine it with 1D-CNN, such as STGCN. Although follow-up works GWN \cite{DBLP:conf/ijcai/WuPLJZ19}, MTGNN \cite{wu2020connecting} and AGCRN \cite{DBLP:conf/nips/0001YL0020} based on DCRNN and STGCN have improved performance, the static perception of graph convolution and the cumulative error of iterative prediction still limits them.

ST-GRAT \cite{park2020st} adopts the Transformer \cite{DBLP:conf/nips/VaswaniSPUJGKP17} structure, which applies the SA to the two dimensions of spatial and temporal to capture the global dependence, but it still uses iterative predictions. GMAN also uses spatio-temporal attention. However, it proposes a novel generative style inference method, which uses embedding to perform encoder-decoder attention to generate future sequences without dynamic decoding. Compared with dynamic decoding methods GMAN has improved the long-term forecasting performance but underperformed in the short-term. Regardless of whether the model is based on GCN or SA, the spatio-temporal dimensions are separated without unified modeling.
\subsubsection{Graph Convolution Networks}
GCN is a convolution operation defined on a non-Euclidean structure by \cite{bruna2013spectral}. It extracts spatial features through transformation and aggregation, and has achieved success in many tasks \cite{qiu2018deepinf,li2019encoding}. However, the time complexity of eigendecomposition of the normalized graph Laplacian matrix is too high, kipf et al. \cite{DBLP:conf/iclr/KipfW17} suggested that graph convolution operation can be further approximated by the K-th order polynomial of Laplacians and improve the calculation speed without loss of accuracy. Subsequent work has focused on solving the smoothing problem, i.e., when the number of GCN layers is stacked too much, the characteristics will tend to be consistent. JKNet \cite{xu2018representation} combines the results of multi-layer GCN through skip connections to enhance the expressive ability of the model. Dropedge \cite{DBLP:conf/iclr/RongHXH20} reduces the effect of over-smoothing by deleting the edges of the input graph. But the effects of these two methods are not obvious. On the other hand, some methods use deeper receptive fields for shallow networks. SGC \cite{wu2019simplifying} attempts to use the K-th power of the graph convolution matrix to capture long distance information in a single neural network layer. PPNP and APPNP \cite{DBLP:conf/iclr/KlicperaBG19} replace the power of the graph convolution matrix with the Personalized PageRank matrix to solve the over-smoothing problem. GDC \cite{DBLP:conf/nips/KlicperaWG19} further extends APPNP by generalizing Personalized PageRank to an arbitrary graph diffusion process.
\subsubsection{Attention Mechanism}
The earliest attention appeared in the Bahdanau et al. encoder-decoder framework \cite{bahdanau2014neural}, they proposed the addictive attention. Later, its variant \cite{DBLP:conf/emnlp/LuongPM15} proposed the most commonly used dot product attention for natural language processing. Vaswani et al. \cite{DBLP:conf/nips/VaswaniSPUJGKP17} then proposed the most famous Transformer architecture, which completely uses the self-attention mechanism to solve tasks (e.g. machine translation \cite{brown1990statistical} and question answering system \cite{ravichandran2002learning}) and achieved superior performance, making the self-attention mechanism no longer just an aid to the encoder-decoder framework. In the next period, Transformer-based works have sprung up in computer vision, speech and signal processing \cite{owens1993signal}, bioinformatics \cite{baxevanis2020bioinformatics}, etc., and have achieved the best results in various tasks. However, self-attention has a quadratic time and memory complexity issue, which hinders model scalability in many settings. To address this problem, Sparse Transformer \cite{child2019generating} and Longformer \cite{beltagy2020longformer} utilize pre-defined fixed local pattern to optimize self-attention; Reformer \cite{DBLP:conf/iclr/KitaevKL20} design a hash-based similarity measurement and Routing Transformer \cite{roy2021efficient} design an online k-means clustering, they are efficiently learn local patterns; Linformer \cite{wang2020linformer} found that the feature map matrix in self-attention has low rank and they map key to a low-rank matrix ($N\to k$) to transform the $N\times N$ calculation to $N\times k$; Performer \cite{choromanski2020rethinking} enables clever mathematical re-writing of the self-attention and no longer calculates the $N\times N$ matrix.
\begin{figure*}[t]
    \centering
    \includegraphics[width=1.0\linewidth]{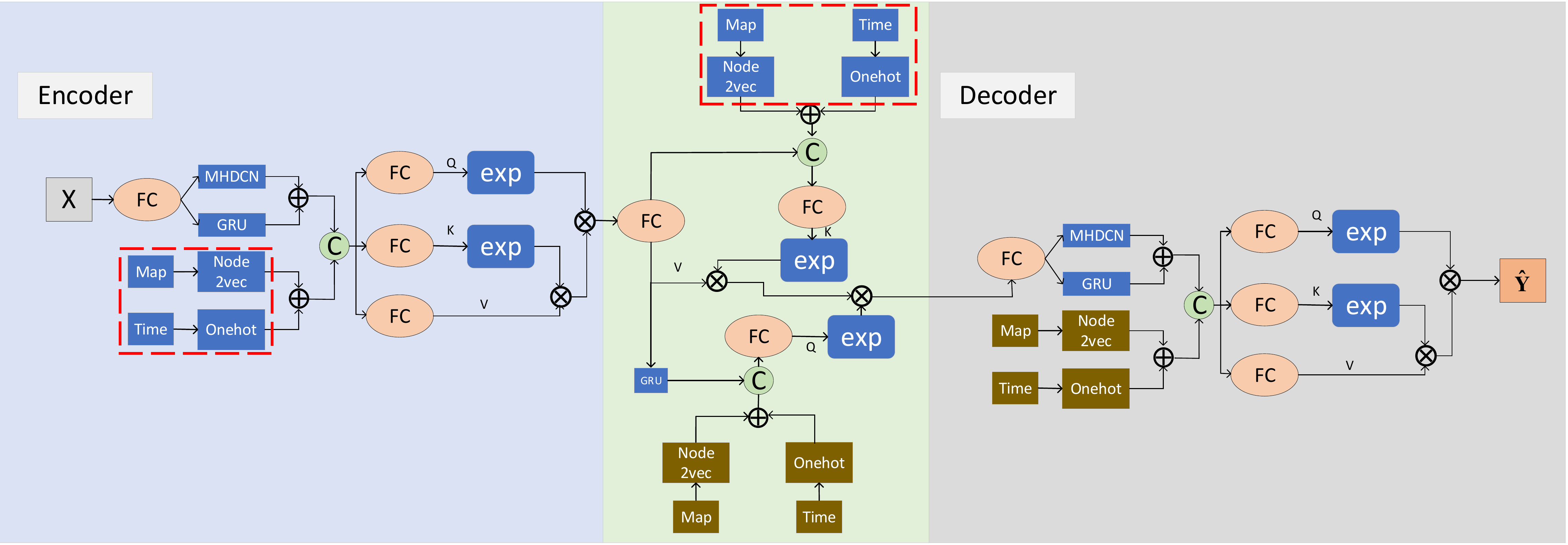}
    \caption{Figure 2 illustrates the structure of our proposed STJLA, which comprises an encoder-decoder framework, and a transform layer between in encoder and decoder converts the historical sequences into the future sequences. $\oplus$, $\otimes$, exp and circle with \textbf{C} indicates add, matrix multiply, exponential and concatenate operation. Besides, the blue \textbf{map} and \textbf{time} in the red box represent the past spatio-temporal information, and the yellow \textbf{map} and \textbf{time} represent the future spatio-temporal information.}
\end{figure*}
\section{Preliminaries}
\subsection{Problem Definition}
The goal of traffic forecasting is to predict the future traffic data given previously observed traffic data from $N$ correlated sensors on the road network. We define the $N$ correlated traffic sensors as a weighted directed graph $\mathcal{G}=(\mathcal{V},\mathcal{E},A)$, where $\mathcal{V}$ is a set of vertices representing traffic sensors and $\mathcal{\vert V\vert} = N$, $\mathcal{E}$ is a set of edges and $A\in\mathbb{R}^{N\times N}$ is a weighted adjacency matrix of $\mathcal{G}$ representing the nodes proximity. Denote the traffic data observed on node $i$ at time $t$ as a graph signal $X_t^i\in\mathbb{R}^{C}$, where $C$ is the number of features of each node (e.g. traffic flow and traffic speed). The traffic forecasting problem aims to learn a function $f$ for forecasting the $T_p$ future graph signals $\mathcal{Y}\in\mathbb{R}^{T_p\times N\times C}$ from the known $T_h$ graph signals $\mathcal{X}\in\mathbb{R}^{T_h\times N\times C}$ and the graph $\mathcal{G}$:
\begin{equation}\nonumber
[X_{1},...,X_{T_h};\mathcal{G}]\mathop{\rightarrow}\limits_{f(\cdot)}[Y_{1},...,Y_{T_p}]
\end{equation}

\subsection{Diffusion Convolution Networks}
\cite{DBLP:conf/iclr/LiYS018} proposed a diffusion convolution network for better suiting the nature of traffic flow. Specifically, \cite{teng2016scalable} proved that the diffusion process of information on graph $\mathcal{G}$ can be formulated as a process of random work with restart probability $\alpha\in[0, 1]$ and transition matrix $D^{-1}_OA$. $D_O=diag(A_1)$ is the out-degree diagonal matrix. The stationary distribution of the diffusion process can be represented as a weighted combination of infinite random walks on the graph and be calculated in closed form:
\begin{equation}
\mathcal{P}=\sum\limits_{k=0}^{\infty}\alpha(1-\alpha)^k(D^{-1}_OA)^k
\end{equation}
where $\mathcal{P}\in\mathbb{R}^{N\times N}$ denotes the stationary distribution of the Markov process of diffusion. In DCRNN, authors used a finite $K$ step truncation of the diffusion process and assigned a trainable weight to each step. Therefore, the step $k$ of uni-directional process of information diffusion in DCRNN can be rewritten as:
\begin{equation}
X_{\star G}\Theta=((D^{-1}_OA)^k+(D^{-1}_IA^T)^k)XW
\end{equation}
where $X\in\mathbb{R}^{N\times F}$ is the input signal, $W\in\mathbb{R}^{F\times F}$ is the parameter for the filter and $D^{-1}_OA$ , $D^{-1}_IA^T$ represent the transition matrices of the diffusion process and the reverse one, respectively.
\subsection{Self-attention}
The attention mechanism first calculates the correlation between each position, and then uses the calculated feature map matrix to aggregate information. Formally, the input
sequence $X\in\mathbb{R}^{N\times F}$ is projected by three matrices $W^Q\in\mathbb{R}^{F\times F}$, $W^K\in\mathbb{R}^{F\times F}$ and $W^V\in\mathbb{R}^{F\times F}$ to corresponding representations $Q$, $K$ and $V$:
\begin{equation}
    \begin{split}
        &Q=XW^Q\\
        &K=XW^K\\
        &V=XW^V\\
    \end{split}
\end{equation}
then the self-attention is computed as follows:
\begin{equation}
    Attention(Q,K,V) = softmax(\frac{QK^T}{\sqrt{F}})V
\end{equation}

From \cite{katharopoulos2020transformers}, we can write self-attention in a more general form by replacing softmax with a similarity function:
\begin{equation}
    Attention(Q,K,V)_i = \frac{\sum_{j=1}^Nsim(Q_i,K_j)V_j}{\sum_{j=1}^Nsim(Q_i,K_j)}
\end{equation}
when we substitute the similarity function with $sim(q,k)=exp(\frac{qk^T}{\sqrt{F}})$, the Equation 5 is s equivalent to equation 4.
\section{Methodology}
\subsection{Model Overview}
Fig. 2 illustrates the structure of our proposed STJLA, which comprises an encoder-decoder framework, and a transform layer between in encoder and decoder converts the historical sequences into the future sequences. The encoder and decoder have the same structure, which extracts the dynamic semantic context from MHDCN and GRU, and extracts the static structure context from node2vec \cite{grover2016node2vec} and one-hot encoding. Finally, the dynamic context and static context are concatenated as the input of the linear attention. These modules are detailed next.
\subsection{Input Layer and Output Layer}
We use a fully connected layer at the top of the model to project the input $\mathcal{X}\in\mathbb{R}^{T_h\times N\times C}$ into a high-dimensional space $\mathcal{X}\in\mathbb{R}^{T_h\times N\times F}$ to enhance the representation power of the model. Similarly, we use a fully connected layer at the bottom of the model to project the output from high-dimensional space $\mathcal{Y}\in\mathbb{R}^{T_h\times N\times F}$ to traffic flow $\mathcal{Y}\in\mathbb{R}^{T_h\times N\times C}$.
\subsection{Static Spatial Structure Context}
Topological structure contains rich spatial information, and now there are many graph representation learning methods that represent nodes with high-dimensional context features through topological structures. We use the node2vec algorithm to extract the context information as a spatial embedding vector $SSC\in\mathbb{R}^{N\times 64}$ from the static spatial structure to make the nodes easy to distinguish. Furthermore, we add a fully connected layer to convert the dimension of $SSC$ into $F$ for persisting with model dimension.
\subsection{Static Temporal Structure Context}
The traffic dataset generally aggregates data into 5-15 minutes window. We assume that the time window is 5 minutes, and each day consists of 288 time slices. We found that the same time slices on different days always have the same trend, such as morning and evening peak. In addition, Mondays of different weeks also have the same trend, but different days in one week always have different trends. Therefore, we use one-hot encoding to encode the position of the time slice in day and week into a vector $TSC\in\mathbb{R}^{T\times 295}$ ($T=T_h$ in encoder and $T=T_p$ in decoder), and in order to be consistent with the model dimension, we use a fully connected layer to project the vector to $TSC\in\mathbb{R}^{T\times F}$. In this way, the static temporal context obtained based on the timeline can capture daily trends and weekly trends.
\subsection{Dynamic Spatial Semantic Context}
The context extracted from the static spatial structure cannot fully reflect spatial correlations between nodes. First, because the pre-defined spatial structure according to the road network may be incomplete or even wrong, the spatial context extracted from the wrong prior knowledge may also be incomplete or even biased. Second, the context of sensors in the road network is not static but changes dynamically. In other words, the traffic flow in the road network does not follow a fixed route. People always have multiple choices, but static spatial context are fixed. The road network not contains the dynamic spatial context. Therefore, we use MHDCN to dynamically learn spatial contexts at different times through semantic information. In addition, the MHDCN enhances the local spatial expression ability of the model because the attention mechanism is locally insensitive.

Similar to methods such as PPNP, MHDCN also increases the receptive field in one convolution operation instead of increasing the number of the convolution operation. However, unlike PPNP uses the adjacency matrix with redundant information after power iteration as the convolution kernel, MHDCN selects neighbors with different hops in the shortest path tree of each node as multi-hop adjacency matrix. Algorithm 1 shows how to select 1 to k hop neighbors of each node. More specifically, inspired by the self-attention mechanism, we use the selected multi-hop adjacency matrix $\mathcal{H}\in\mathbb{R}^{k\times N\times N}$ for multi-head operation to obtain information from the space of different hop subgraphs. MHDCN for the input $X_t\in\mathbb{R}^{N\times F}$ is:
\begin{equation}
    \begin{split}
        &MHDCN(X_t)=Concat(head_1,...,head_k)W^D\\
        &where\quad head_i=(D^{-1}_O\mathcal{H}_i+D^{-1}_I\mathcal{H}_i^T)(X_tW^X)
    \end{split}
\end{equation}
where the projections are parameter matrices $W^X\in\mathbb{R}^{F\times \frac{F}{k}}$ and $W^D\in\mathbb{R}^{F\times F}$. $\mathcal{H}_i$ indicates the outflow matrix and $\mathcal{H}_i^T$ indicates the inflow matrix. Besides, MHDCN normalizes the inflow and outflow matrix through their respective degree matrices $D^{-1}_I$ and $D^{-1}_O$.
\begin{algorithm}
\caption{Find the 1 to k hop neighbors of each node.}
\KwData{Road network $\mathcal{G}=(\mathcal{V},\mathcal{E},A)$;}
Initialize a shortest path matrix $\mathcal{S}\in\mathbb{R}^{N\times N}$ with zero;
\For{$i\leftarrow 1$ \KwTo $N$}{
    \For{$j\leftarrow 1$ \KwTo $N$}{
        Compute the shortest distance $d$ between node $i$ and node $j$ based on $\mathcal{G}$, and $\mathcal{S}[i,j]=d$;
    }
}
\For{$i\leftarrow 1$ \KwTo $k$}{
    Generate a adjacency matrix $\mathcal{H}_{i}\in\mathbb{R}^{N\times N}$\\
    \For{$j\leftarrow 1$ \KwTo $N$}{
        \For{$l\leftarrow 1$ \KwTo $N$}{
            $\mathcal{H}_{i}[j,l]=\begin{cases} 1& , \mathcal{S}[j,l]=i\\0& , otherwise \end{cases}$
        }
    }
}
\KwResult{The 1 to k hop adjacency matrix $\{\mathcal{H}_{1:k}\}$}
\end{algorithm}
\subsection{Dynamic Temporal Semantic Context}
The static temporal structure context based on the position of the time sequence captures daily and weekly traffic trends, but cannot represent hourly traffic changes. Therefore, we propose a RNN-based model to extract dynamic temporal semantic context to further capture fine-grained hourly traffic changes. On the other hand, the spatio-temporal joint attention mechanism is not auto-regressive, i.e., the forecast results are not superior to the long-term in the short-term. Although the causal spatio-temporal joint attention mechanism satisfies auto-regressive, it performs poorly in the joint graph of long sequences. So we choose RNN to realize the auto-regressive of our model. In this paper, we utilize GRU to implement a recurrent network. GRU is a variant of RNN. Each unit will output a hidden state as part of the inputs of the next unit. Information always flows from one unit to the next unit in the GRU, and historical units will not get information about future units. Thus GRU is an auto-regressive model capable of capturing temporal correlations in sequence. Besides, GRU is proposed to address the problem of vanishing or exploding of gradient in RNN, so that it could better learn long-term temporal correlations. Formally:
\begin{equation}
    \begin{split}
        &R_t=\sigma(X_tW_{xr}+H_{t-1}W_{hr}+b_r)\\
        &U_t=\sigma(X_tW_{xu}+H_{t-1}W_{hu}+b_u)\\
        &\tilde{H}_t=tanh(X_tW_{xh}+(R_t\odot H_{t-1})W_{hh}+b_h)\\
        &H_t=U_t\odot H_{t-1}+(1-U_t)\odot \tilde{H}_t\\
    \end{split}
\end{equation}
where $X_t\in\mathbb{R}^{N\times F}$ indicates the input of GRU at time slice $t$ and $H_{t-1}\in\mathbb{R}^{N\times F}$ indicates the output hidden state at time slice $t-1$, which contain historical information. More specifically, $H_0$ is constructed by zero vectors and the last encoder hidden state in encoder and decoder, respectively. $W_{xr}$, $W_{hr}$, $W_{xu}$, $W_{hu}$, $W_{xh}$ and $W_{hh}\in\mathbb{R}^{F\times F}$ are learnable parameters in STJLA. $b_r$, $b_u$ and $b_h\in\mathbb{R}^{F}$ are bias in STJLA. The $\sigma$ represents sigmoid activation function, which restricts the value of all elements in reset gate $R_t\in\mathbb{R}^{N\times F}$ and update gate $U_t\in\mathbb{R}^{N\times F}$ between 0 and 1 to capture short and long-term temporal dependencies, respectively.
\subsection{Spatio-Temporal Joint Linear Attention}
In previous works, GMAN and ST-GRAT used the self-attention mechanism in the spatial or temporal dimension to extract spatial and temporal features separately, and then merge the temporal and spatial features. In other words, spatial attention utilizes self-attention to each time slice $X_t\in\mathbb{R}^{N\times F}$ to extract spatial correlations in this time slice. Temporal attention utilizes self-attention to each node $X^i\in\mathbb{R}^{T\times F}$ to capture temporal correlations of uni-node. We argue that the use of temporal and spatial attention can capture the spatial dependence between current node and history nodes, because the node in future should not obtain all the information in the past, even the information is compressed to store in one historical state. Therefore, we believe that the temporal and spatial dimensions should be combined, i.e., the node information of all time slices should be merged into a spatio-temporal joint graph. The spatio-temporal joint graph saves traffic information for all nodes. Therefore, we use the self-attention mechanism for the graph can capture all spatio-temporal correlations instead of individual temporal or spatial correlations. We reshape the input $\mathcal{X}\in\mathbb{R}^{T\times N\times F}$ into $\tilde{X}\in\mathbb{R}^{TN\times F}$ to achieve spatio-temporal joint attention. Besides, we use the multi-head attention (MHA) to jointly attend to information from different representation subspaces. Formally:
\begin{equation}
    \begin{split}
        &MHA(\tilde{X}) = Concat(head_1,...,head_K)W^O\\
        &where\quad head_i=Attention(\tilde{X}W^Q,\tilde{X}W^K,\tilde{X}W^V)\\
    \end{split}
\end{equation}
where $K$ indicates the number of heads in MHA. The projections are parameter matrices $W^Q\in\mathbb{R}^{F\times \frac{F}{K}}$, $W^K\in\mathbb{R}^{F\times \frac{F}{K}}$, $W^V\in\mathbb{R}^{F\times \frac{F}{K}}$ and $W^O\in\mathbb{R}^{F\times F}$.

The disadvantage of the self-attention mechanism is that the quadratic problem will greatly increase the consumption of computing resources. The models that use temporal attention and spatial attention are very large and slow and the higher complexity of the spatio-temporal joint attention mechanism makes the model impossible to train (time complexity of temporal, spatial and spatio-temporal joint attention is $O(NT^2)$, $O(TN^2)$ and $O(T^2N^2)$, respectively). Inspired by \cite{katharopoulos2020transformers}, we note that only need to ensure the non-negativity of $sim(\cdot)$ to define the attention function in equation 5. This includes all kernels $sim(x,y):\mathbb{R}^{2\times F}\to \mathbb{R}_+$. The kernel with a positive feature representation $\phi(x)$ can rewrite equation 5 as follow:
\begin{equation}
    Attention(Q,K,V)_i=\frac{\sum_{j=1}^N\phi(Q_i)\phi(K_j^T)V_j}{\sum_{j=1}^N\phi(Q_i)\phi(K_j^T)}
\end{equation}
and we can further optimize it using the associative property of matrix multiplication, formally:
\begin{equation}
    Attention(Q,K,V)_i=\frac{\phi(Q_i)\sum_{j=1}^N\phi(K_j^T)V_j}{\phi(Q_i)\sum_{j=1}^N\phi(K_j^T)}
\end{equation}

For our experiments, we employ a positive similar function as defined below:
\begin{equation}
    \phi(x) = exp(x)
\end{equation}
where $exp(\cdot)$ operation derive from Performer \cite{choromanski2020rethinking}.

By changing the kernel function, we optimize the time complexity of the self-attention mechanism to linear, and the time complexity of spatio-temporal joint linear attention is $O(TN)$.
\subsection{Transform Layer}
We add a Transform layer between the encoder and the decoder to convert historical information into a future state. In order to avoid using static context based linear attention to generate the future sequence, we first use the last time slice state of the encoder as the start token and the hidden state of the GRU to dynamic decode the future sequence, i.e., $H_{-1}=X_{T_h}\in\mathbb{R}^{N\times F}$ and $Y_t=H_{t-1}$ in GRU. Then we concatenate the generated future sequence $\mathcal{Y}\in\mathbb{R}^{T_p\times N\times F}$ from GRU with future static context as the query of linear attention and concatenate the historical sequence from encoder with historical static context as the key of linear attention to generate the future sequence which obtains global historical information and is the input of the decoder.
\subsection{Loss Function}
For our experiments, we choose L1 loss as our training objective function and optimize the loss for multi-step prediction via back-propagation and Adam optimizer \cite{DBLP:journals/corr/KingmaB14}. Formally:
\begin{equation}
    \mathcal{L}(\Theta)=\sum\limits_{t=1}^{T_p}\sum\limits_{i=1}^N\vert X^i_t-Y_t^i\vert
\end{equation}
where $\Theta$ indicates all the learnable parameters in our model. $X^i_t$ represents the traffic information at the time $t$ of node $i$ and $Y^i_t$ represents the predicted value at the time $t$ of node $i$, respectively. The training procedure of STJLA is summarized in Algorithm 2.
\begin{algorithm}
\caption{Training procedure of STJLA.}
\KwData{Road network graph $\mathcal{G}=(\mathcal{V},\mathcal{E},A)$;\\Time steps $TS$ of train set;\\Train data of all observing stations $\mathcal{TD}\in\mathbb{R}^{TS\times N}$;\\All hyperparameters;}
Employ node2vec and onehot encoding to get the $\mathcal{SSC}\in\mathbb{R}^{N\times 64}$ and $\mathcal{TSC}\in\mathbb{R}^{TS\times 295}$;\\
\For{$t\leftarrow 1$ \KwTo $TS-24$}{
    Append $\mathcal{TD}_{t:t+12}$ to input sample set $\mathcal{X}$;\\
    Append $\mathcal{TD}_{t+12:t+24}$ to label sample set $\mathcal{Y}$;\\
    Append $\mathcal{TSC}_{t:t+24}$ to context sample set $\mathcal{C}$;\\
}
Initialize all learnable parameters $\Theta$ in STJLA;\\
\Repeat{met model stop criteria}{
Randomly select a batch of input sample $\mathcal{X}_{bs}$ form $\mathcal{X}$;\\
Randomly select a batch of label sample $\mathcal{Y}_{bs}$ form $\mathcal{Y}$;\\
Randomly select a batch of context sample $\mathcal{C}_{bs}$ form $\mathcal{C}$;\\
Optimize $\Theta$ by minimizing the objective function
Eq.12 with $\mathcal{X}_{bs}$, $\mathcal{Y}_{bs}$, $\mathcal{C}_{bs}$ and $\mathcal{SSC}$ as input;
}
\KwResult{Learned STJLA model.}
\end{algorithm}
\begin{table}[htbp]
\centering
\caption{Detailed statistics of England and PEMSD7.}
\begin{tabular}{llllll}
\toprule
Dataset & \#Nodes & \#Edges & \#Time Steps & \#Mean & \#Std\\
\midrule
England & 249 & 1925 & 35040 & 427 & 382\\
PEMSD7 & 228 & 2036 & 12672 & 59 & 13\\
\bottomrule
\end{tabular}
\end{table}
\section{Experiments}
\subsection{Datasets}
We test STJLA on two public traffic datasets, England \cite{zhang2021graph} and PEMSD7 \cite{DBLP:conf/ijcai/YuYZ18}. England leverage 249 sensors to record traffic flow data on the England highways from January 1st, 2014 to December 31th, 2014, and PEMSD7 leverage 228 sensors to record traffic speed data on the Ventura area in America from May 1st 2012 to June 31th 2012. England aggregate traffic flow observations into 15 minutes windows and PEMSD7 aggregate traffic speed observations into 5 minutes windows, respectively. Besides, we apply Z-Score normalization to normalize data before in the model. Both the datasets are split in chronological order with 70\% for training, 10\% for validation, and 20\% for testing. Table \uppercase\expandafter{\romannumeral1} shows the detailed statistics of the datasets.
\subsection{Baselines}
We compare our model with the following models:
\begin{itemize}
\item VAR \cite{lu2016integrating}, vector Auto-Regression, which captures the pairwise relationships among multiple time series and we set order of 12 in experiments.
\item SVR \cite{wu2004travel}, support vector regression model. We use support vector with rbf kernel to iteratively perform multi-step predictions.
\item LSTM \cite{hochreiter1997long}, long short-term memory network, which is a variant of recurrent neural network. Compared with ordinary rnn, LSTM effectively prevents the gradient vanish and capture the long range temporal dependence.
\item DCRNN \cite{DBLP:conf/iclr/LiYS018}. Diffusion convolution recurrent neural network, which replaces the linear operation in the GRU with a diffusion graph convolution network to extract the spatial dependence instead of simply doing multiple univariate time series forecasting like LSTM or traditional methods. Besides, DCRNN uses a encoder-decoder architecture for multi-step forecasting and uses schedule sampling to speed up convergence.
\item STGCN \cite{DBLP:conf/ijcai/YuYZ18}. Spatial-temporal graph convolution network uses graph convolution to capture spatial dependence, and uses gated linear unit (GLU) \cite{dauphin2017language} to extract temporal dependence, which greatly accelerates the training speed compared with using RNN.
\item GWN \cite{DBLP:conf/ijcai/WuPLJZ19}. A advanced model based on STGCN uses the self-adaptive adjacency matrix learnt by data-driven manner in graph convolution to capture spatial dependence instead of using the adjacency matrix obtained from the road network. It further uses Gated TCN instead of GLU to capture temporal dependence.
\item AGCRN \cite{DBLP:conf/nips/0001YL0020}. A advanced model based on DCRNN uses the self-adaptive adjacency matrix learnt by data-driven manner in graph convolution to capture spatial dependence. Besides, it proposed a novel node adaptive parameter learning method to reduce parameters of graph convolution.
\item MTGNN \cite{wu2020connecting}. MTGNN no longer uses any prior knowledge for graph convolution. Furthermore, it uses Mixhop and Dilated inception to capture spatio-temporal dependencies. Compared with GWN, it obtains multi-scale spatio-temporal information.
\item ST-GRAT \cite{park2020st}, which adopts the Transformer architecture and adds a novel spatial attention mechanism that uses additional sentinels and diffusion graph matrix to the model to extract the spatial dependence.
\item GMAN \cite{zheng2020gman}. A Graph Multi-Attention Network, which adopts an encoder-decoder architecture, where both the encoder and the decoder consist of multiple spatio-temporal attention blocks to capture spatio-temporal dependencies. In addition, GMAN design a novel generation inference methods to avoid dynamic decoding in Transformer.
\end{itemize}
\subsection{Metrics}
We use three methods: mean absolute error (MAE), root mean square error (RMSE), and mean absolute percentage error (MAPE) to evaluate the performance for all experiments on the two datasets of PEMSD7 and England. The three metric formulated as follows:
\begin{equation}
    MAE = \sum\limits_{t=1}^T\sum\limits_{i=1}^N\vert X_t^i-Y_t^i\vert
\end{equation}
\begin{equation}
    RMSE = \sum\limits_{t=1}^T\sum\limits_{i=1}^N \sqrt{(X_t^i-Y_t^i)^2}
\end{equation}
\begin{equation}
    MAPE = \sum\limits_{t=1}^T\sum\limits_{i=1}^N\vert \frac{X_t^i-Y_t^i}{X_t^i}\vert
\end{equation}
\subsection{Experimental Settings}
Our experiments are conducted under a computer environment with one Intel (R) Xeon (R) Gold 6132 CPU and one Tesla-V100 GPU card. We train our model by using Adam optimizer to minimize the L1 loss of 40 and 8 epochs on England and PEMSD7 dataset with the batch size 16 and the initial learning rate 1e-3. Furthermore, the learning rate will be reduced to $\frac{1}{10}$ of the original in [5,6,7] and [25,35] epoch on PEMSD7 and England dataset, respectively. Besides, the number of dimension $F$ in our model is 128. Totally, there are 4 hyperparameters in STJLA, the number of linear attention heads $K$, the dimension of each head $d$, the number of multi-head diffusion convolution network heads $k$ and the GRU layers $L$. We tune these parameters on the validation set, and observe the best performance on the setting is $K=8$, $d=16$, $k=8$ and $L=2$.

For baseline experiments, all deep learning models except LSTM and ST-GRAT use the open source code and the optimal parameters given in the paper. We reproduced ST-GRAT according to the paper and achieved results equivalent to the paper, and realized the LSTM with 128 hidden units and three layers. Besides, we use the statsmodels and sklearn packages in python to implement the VAR and SVR.
\begin{table*}[t]
\centering
\caption{Traffic forecasting performance comparison of STJLA and baselines on England dataset.}
\resizebox{1.0\linewidth}{!}{\begin{tabular}{lllll|lll|lll|lll} 
\toprule
 & \multirow{2}{*}{Model} & \multicolumn{3}{c|}{45 min} & \multicolumn{3}{c|}{90 min} & \multicolumn{3}{c|}{3 hour} & \multicolumn{3}{c}{Average}\\
\cmidrule{3-14}
    &  & MAE & RMSE & MAPE & MAE & RMSE & MAPE & MAE & RMSE & MAPE & MAE & RMSE & MAPE \\
\midrule
\multirow{13}{*}{England}
& VAR & 50.14 & 78.72 & 36.46\% & 82.66 & 122.82 & 60.57\% & 105.21 & 150.72 & 83.10\% & 69.28 & 109.32 & 51.81\%\\
& SVR & 54.42 & 85.88 & 53.23\% & 97.02 & 152.44 & 70.12\% & 139.35 & 223.89 & 74.75$\%$ & 82.49 & 142.73 & 46.48$\%$\\
& LSTM & 58.09 & 97.87 & 28.85 & 97.56 & 159.68 & 48.61 & 131.22 & 208.38 & 70.55$\%$ & 82.23 & 143.85 & 41.75$\%$\\
\cmidrule{2-14}
& DCRNN & 34.87 & 59.97 & 25.14\% & 41.02 & 71.92 & 30.10\% & 46.45 & 83.21 & 34.59$\%$ & 38.34 & 67.76 & 27.92$\%$\\
& STGCN & 34.11 & 59.13 & 25.92$\%$ & 48.30 & 83.83 & 34.18$\%$ & 65.25 & 117.96 & 42.02$\%$ & 43.47 & 79.86 & 30.48$\%$\\
\cmidrule{2-14}
& GWN & 31.94 & 56.94 & 21.07$\%$ & 38.75 & 66.83 & 28.08$\%$ & 43.06 & 74.99 & \underline{33.14}$\%$ & \underline{35.40} & 61.48 & 24.70$\%$\\
& AGCRN & 33.43 & 57.94 & 26.59$\%$ & 39.25 & 68.21 & 31.31$\%$ & 44.20 & 76.99 & 34.97$\%$ & 36.72 & 76.99 & 34.97$\%$\\
& MTGNN & 32.47 & 56.45 & 25.11$\%$ & 38.22 & \underline{65.71} & 31.79$\%$ & 42.69 & 72.88 & 36.23$\%$ & 35.44 & \underline{60.93} & 28.48$\%$\\
\cmidrule{2-14}
& GMAN & 33.85 & 58.69 & 29.28\% & \underline{37.87} & 65.77 & 32.73\% & \underline{41.20} & \underline{70.94} & 36.15$\%$ & 36.15 & 62.85 & 31.31$\%$\\
& ST-GRAT & \underline{31.81} & \underline{56.43} & \underline{20.79}\% & 41.14 & 73.06 & \underline{27.76}\% & 50.27 & 94.15 & 33.39$\%$ & 37.19 & 68.95 & \underline{24.34}$\%$\\
\cmidrule[0.6pt]{2-14}
& STJLA & \textbf{29.79} & \textbf{53.32} & \textbf{20.74}$\%$ & \textbf{34.28}  & \textbf{61.00} & \textbf{27.44}$\%$ & \textbf{37.49} & \textbf{66.11} & \textbf{31.00}$\%$ & \textbf{31.92} & \textbf{57.20} & \textbf{23.55}$\%$\\
\bottomrule
\end{tabular}}
\end{table*}
\begin{table*}[t]
\centering
\caption{Traffic forecasting performance comparison of STJLA and baselines on PEMSD7 dataset.}
\resizebox{1.0\linewidth}{!}{\begin{tabular}{lllll|lll|lll|lll} 
\toprule
 & \multirow{2}{*}{Model} & \multicolumn{3}{c|}{15 min} & \multicolumn{3}{c|}{30 min} & \multicolumn{3}{c|}{1 hour} & \multicolumn{3}{c}{Average}\\
\cmidrule{3-14}
    &  & MAE & RMSE & MAPE & MAE & RMSE & MAPE & MAE & RMSE & MAPE & MAE & RMSE & MAPE \\
\midrule
\multirow{13}{*}{PEMSD7}
& VAR & 3.46 & 5.07 & 7.91\% & 4.63 & 6.84 & 10.82\% & 5.55 & 8.18 & 13.15\% & 4.38 & 6.66 & 10.22\%\\
& SVR & 2.42 & 4.49 & 5.79$\%$ & 3.35 & 6.37 & 8.28$\%$ & 4.72 & 8.81 & 11.98$\%$ & 3.34 & 6.55 & 8.28$\%$\\
& LSTM & 2.32 & 4.45 & 5.41$\%$ & 3.27 & 6.43 & 8.08$\%$ & 4.63 & 8.84 & 12.35$\%$ & 3.25 & 6.59 & 8.18$\%$\\
\cmidrule{2-14}
& DCRNN & 2.24 & 4.20 & 5.34$\%$ & 2.97 & 5.86 & 7.45$\%$ & 3.78 & 7.47 & 9.89$\%$ & 2.89 & 5.83 & 7.27$\%$\\
& STGCN & 2.23 & 4.13 & 5.25$\%$ & 3.00 & 5.78 & 7.31$\%$ & 4.00 & 7.73 & 9.90$\%$ & 2.95 & 5.87 & 7.16$\%$\\
\cmidrule{2-14}
& GWN & 2.12 & 4.06 & 4.96$\%$ & \underline{2.72} & 5.42 & 6.74$\%$ & 3.25 & \underline{6.55} & \underline{8.30}$\%$ & \underline{2.60} & \underline{5.35} & 6.43$\%$\\
& AGCRN & 2.27 & 4.28 & 5.53$\%$ & 2.76 & \underline{5.41} & 6.91$\%$ & 3.32 & 6.64 & 8.52$\%$ & 2.70 & 5.39 & 6.77$\%$\\
& MTGNN & 2.13 & \underline{4.05} & 5.01$\%$ & 2.74 & 5.52 & \underline{6.71}$\%$ & 3.30 & 6.66 & 8.26$\%$ & 2.66 & 5.55 & \underline{6.42}$\%$\\
\cmidrule{2-14}
& GMAN & 2.35 & 4.63 & 5.78$\%$ & 2.76 & 5.63 & 6.98$\%$ & \underline{3.20} & 6.56 & \underline{8.25}$\%$ & 2.72 & 5.57 & 6.86$\%$\\
& ST-GRAT & \underline{2.10} & 4.09 & \underline{4.94}$\%$ & 2.77 & 5.66 & 6.92$\%$ & 3.52 & 7.22 & 9.25$\%$ & 2.69 & 5.64 & 6.74$\%$\\
\cmidrule[0.6pt]{2-14}
& STJLA & \textbf{2.07} & \textbf{3.95} & \textbf{4.88}$\%$ & \textbf{2.64}  & \textbf{5.27} & \textbf{6.59}$\%$ & \textbf{3.13} & \textbf{6.31} & \textbf{8.15}$\%$ & \textbf{2.52} & \textbf{5.15} & \textbf{6.32}$\%$\\
\bottomrule
\end{tabular}}
\end{table*}
\subsection{Forecasting Performance Comparison}
Table \uppercase\expandafter{\romannumeral2} compares the performance of STJLA and baselines for 1 hour (4 steps), 2 hours (8 steps) and 3 hours (12 steps) ahead traffic flow prediction on England dataset and Table \uppercase\expandafter{\romannumeral3} compares the performance of STJLA and baselines for 15 minutes (3 steps), 30 minutes (6 steps) and 60 minutes (12 steps) ahead traffic speed prediction on PEMSD7 dataset.

We observe these following phenomenon in Table \uppercase\expandafter{\romannumeral2} and Table \uppercase\expandafter{\romannumeral3}: 1) In the absence of high-order manual features, deep learning method LSTM has achieved better performance than machine learning and traditional statistical method SVR and VAR on univariate prediction tasks. 2) The results of STGCN and DCRNN show that deep learning models that consider the spatial dependence will achieve better results in traffic forecasting tasks. 3) Compared with the pre-defined adjacency matrix based DCRNN and STGCN, the self-adaptive adjacency matrix based GCN models (e.g. GWN) and self-attention models (e.g. GMAN) have a global spatial receptive field, so that they can obtain more historical spatial information. 4) In the model with the global receptive field, the GMAN and AGCRN models that remove the inductive bias of locality are slightly less effective than the models MTGNN, GWN and STGRAT that use the inductive bias. 5) Compared with other dynamic iterative decoding models, the generative inference model GMAN has achieved better results in long-term prediction but is not good at short-term prediction. 6) The model based on the attention mechanism is more suitable for large datasets and the model based on graph convolution is more suitable for small datasets.

STJLA uses a spatio-temporal joint graph to learn correlations between all nodes. Besides, STJLA uses dynamic semantic context to enhance local representations to help the model train better because performing using locality inductive bias better than not using. Therefore, we observe STJLA has achieved the best results on all long short-term prediction tasks in the two datasets. STJLA has achieved a 9.83\% and 3.08\% MAE improvement of average value in the England and PEMSD7 datasets compared with the best baseline, respectively.
\begin{table}[htbp]
    \centering
    \caption{Ablation study results of STJLA with the England dataset.}
    \resizebox{1.0\linewidth}{!}{\begin{tabular}{c|c|ccccccc}
    \toprule
        T & Metric & Basic & +DTC & +DSC & +EDT & +STC & +SSC & STJLA\\
    \midrule
        \multirow{3}{*}{1 hour} & MAE & 90.05 & 45.24 & 38.25 & 37.50 & 34.74 & 33.23 & 29.79\\
        & RMSE & 146.02 & 74.82 & 64.06 & 63.25 & 59.47 & 57.44 & 53.32\\
        & MAPE (\%) & 47.82 & 26.63 & 23.59 & 24.55 & 23.73 & 22.85 & 20.74\\
    \midrule
        \multirow{3}{*}{2 hour} & MAE & 90.05 & 62.57 & 48.25 & 44.20 & 41.33 & 38.87 & 34.28\\
        & RMSE & 146.02 & 107.86 & 80.64 & 74.41 & 71.66 & 66.24 & 61.00\\
        & MAPE (\%) & 47.80 & 35.13 & 30.69 & 30.20 & 29.69 & 28.94 & 27.44\\
    \midrule
        \multirow{3}{*}{3 hour} & MAE & 89.98 & 79.06 & 57.56 & 51.94 & 47.047 & 43.87 & 37.49\\
        & RMSE & 145.96 & 143.80 & 97.83 & 89.67 & 83.60 & 73.78 & 66.11\\
        & MAPE (\%) & 47.72 & 43.36 & 36.21 & 34.74 & 34.61 & 33.71 & 31.00\\
    \midrule
        \multirow{3}{*}{Average} & MAE & 90.05 & 56.07 & 36.37 & 41.52 & 38.43 & 35.82 & 31.92\\
        & RMSE & 146.02 & 96.06 & 75.72 & 71.20 & 67.32 & 62.47 & 57.20\\
        & MAPE (\%) & 47.80 & 31.60 & 27.41 & 27.33 & 26.67 & 25.97 & 23.55\\
    \bottomrule
    \end{tabular}}
\end{table}
\begin{figure*}[htbp]
\centering
    \begin{subfigure}{0.33\linewidth}
    \includegraphics[width=\linewidth]{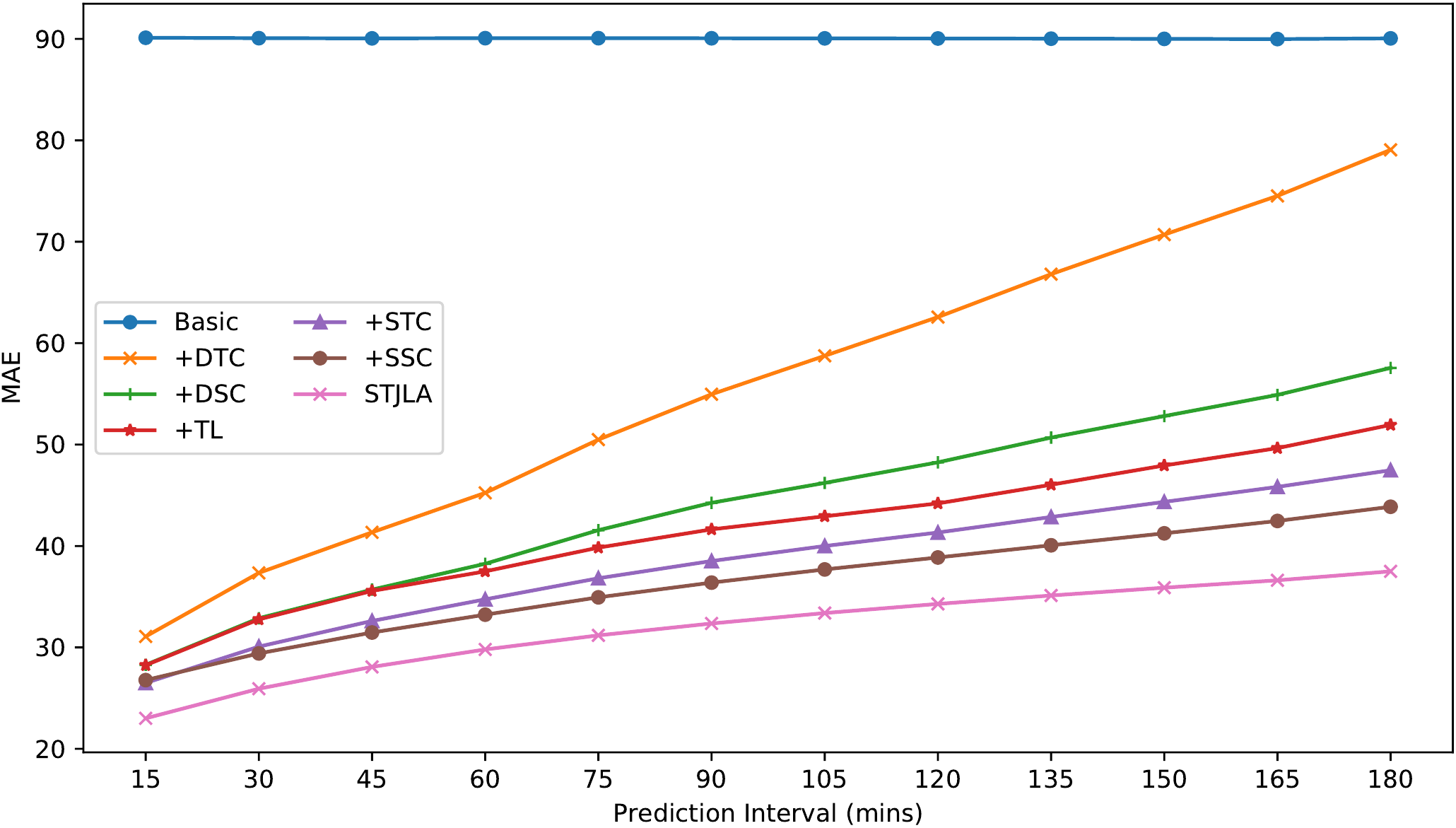}
    \caption{}
  \end{subfigure}%
    \hfill
  \begin{subfigure}{0.33\linewidth}
    \includegraphics[width=\linewidth]{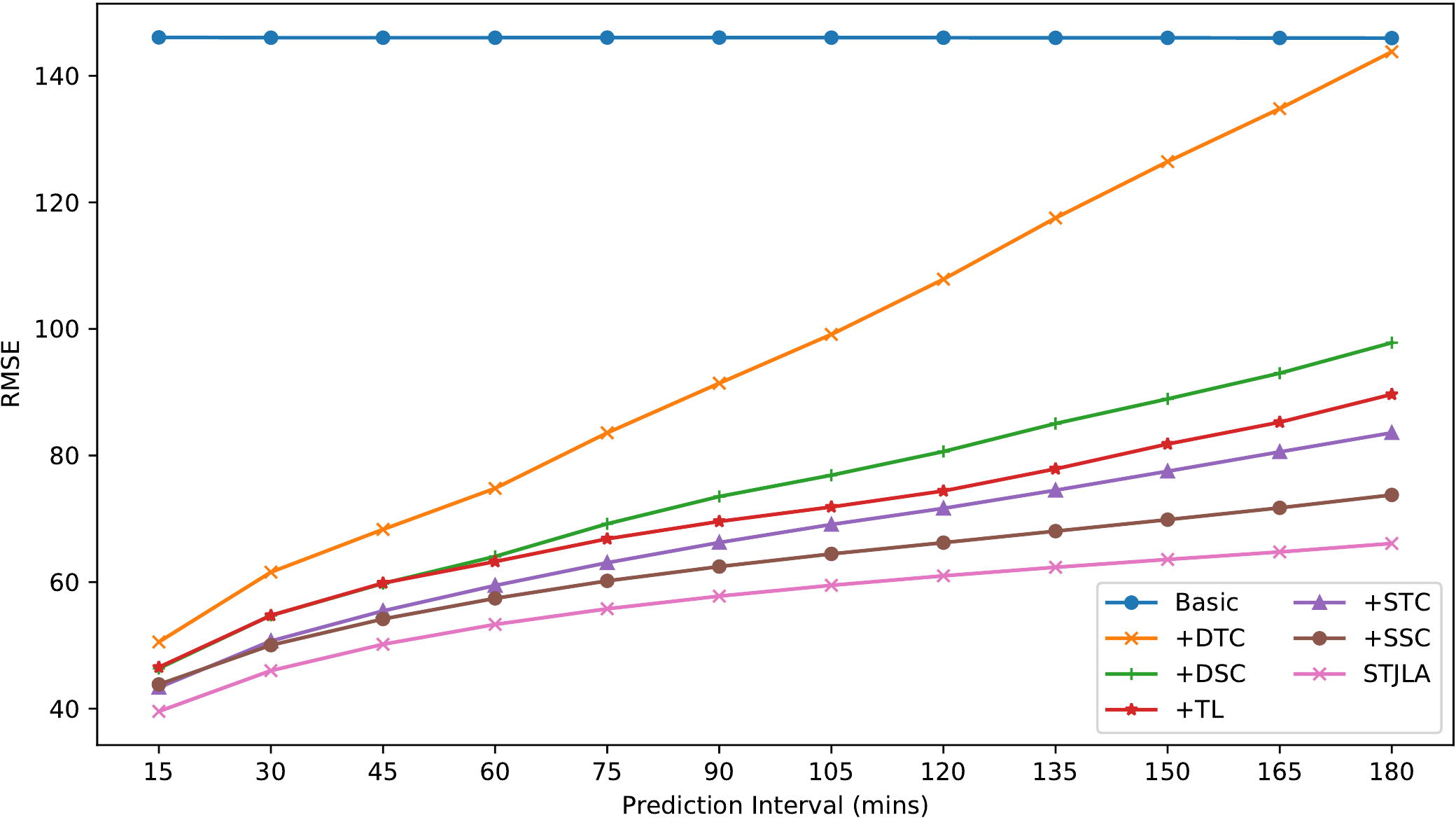}
    \caption{}
  \end{subfigure}%
    \hfill
  \begin{subfigure}{0.33\linewidth}
    \includegraphics[width=\linewidth]{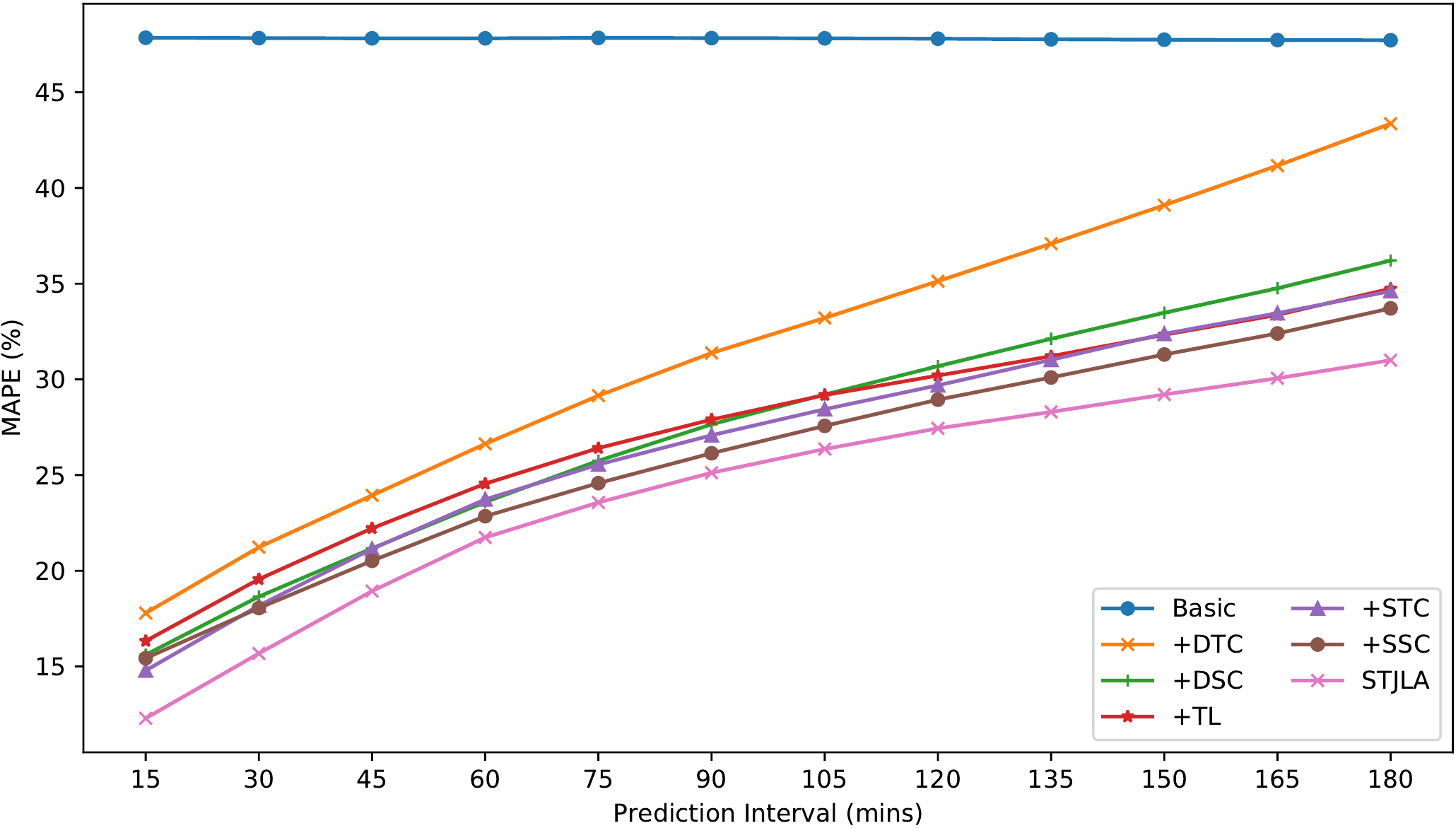}
    \caption{}
  \end{subfigure}%
  
  \caption{Curves composed of the results of each time slice of the ablation study on England dataset.}
\end{figure*}
\subsection{Ablation Study}
We design the following variants of STJLA to verify the effectiveness of each part we proposed in model:
\begin{itemize}
    \item Basic: STJLA is no longer equipped with static contexts, dynamic contexts and the transform layer. Moreover, the input of the linear attention is no longer based on the traffic information of the joint spatio-temporal graph, i.e., the input has changed from $\tilde{X}\in\mathbb{R}^{TN\times F}$ to $\tilde{X}\in\mathbb{R}^{T\times N\times F}$.
    \item +DTC: This model equips Basic with dynamic temporal context.
    \item +DSC: This model equips +DTC with dynamic spatial context.
    \item +TL: This model adds a transform layer on the basis of +DSC.
    \item +STC: This model equips +EDT with static temporal context.
    \item +SSC: This model equips +STC with static spatial context.
    \item STJLA: Compared with the +SSC model, STJLA uses traffic information based on the spatio-temporal joint graph.
\end{itemize}
All the variants have the same settings as STJLA, except the differences mentioned above. Table \uppercase\expandafter{\romannumeral4} shows the prediction results for each hour and the average value over next three hours, while Fig. 3 shows the prediction results for each time slice over the next three hours on the England dataset. First, we found that the basic model cannot extract temporal correlations, which result in each time slice have same error. The +DTC model adds a dynamic temporal semantic context to the basic model so that the +DTC can obtain time sequence relative position information, so the error is no longer a straight line but a rising curve. At the same time, the +DTC model not only provides relative position information but also greatly improves the performance of the model, like the +DSC. In addition, the results of +TL prove the effectiveness of the encoder-decoder architecture and the transform layer we designed, especially for long-term forecasting. Moreover, the models +STC and +SSC, after adding static structure context, have achieved better performance. Finally, by comparing the performance of STJLA and +SSC, we can see that the use of spatio-temporal joint graphs is very important for spatio-temporal forecasting tasks.
\begin{figure}[htbp]
\centering
    \begin{subfigure}{0.48\linewidth}
    \includegraphics[width=\linewidth]{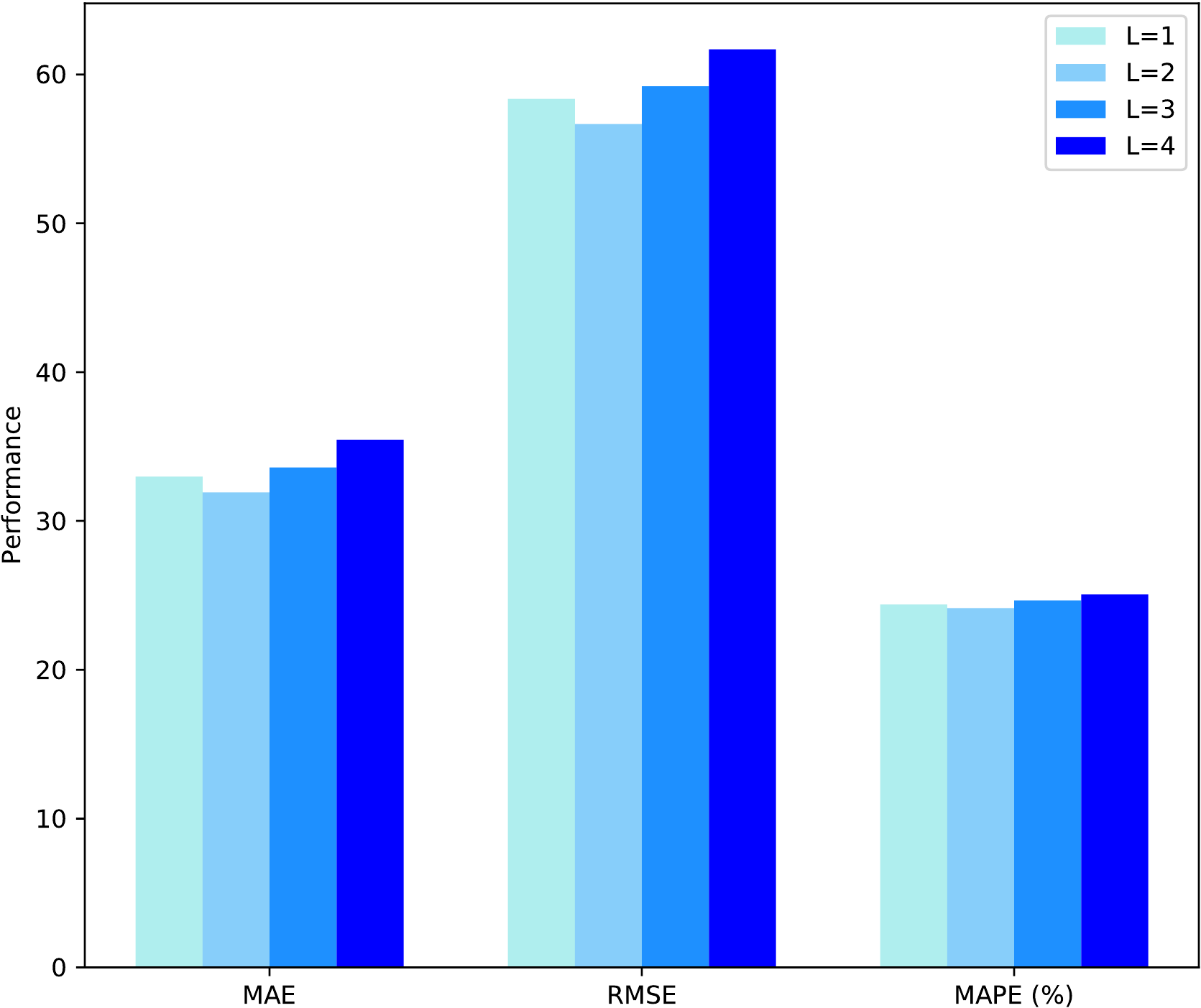}
    \caption{GRU layers.}
  \end{subfigure}%
  \hfill
  \begin{subfigure}{0.48\linewidth}
    \includegraphics[width=\linewidth]{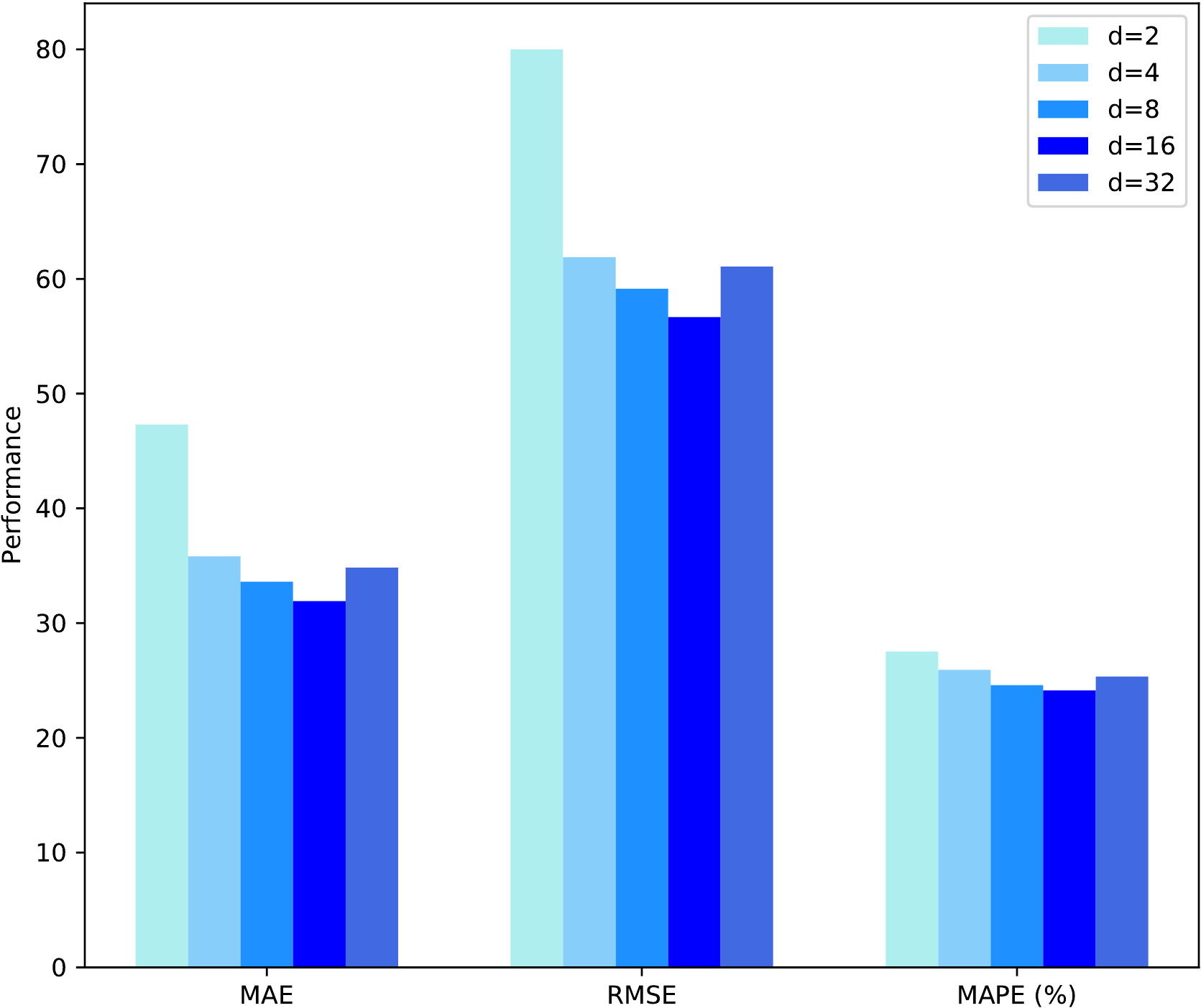}
    \caption{Dimensions of each head.}
  \end{subfigure}%

  \begin{subfigure}{0.48\linewidth}
    \includegraphics[width=\linewidth]{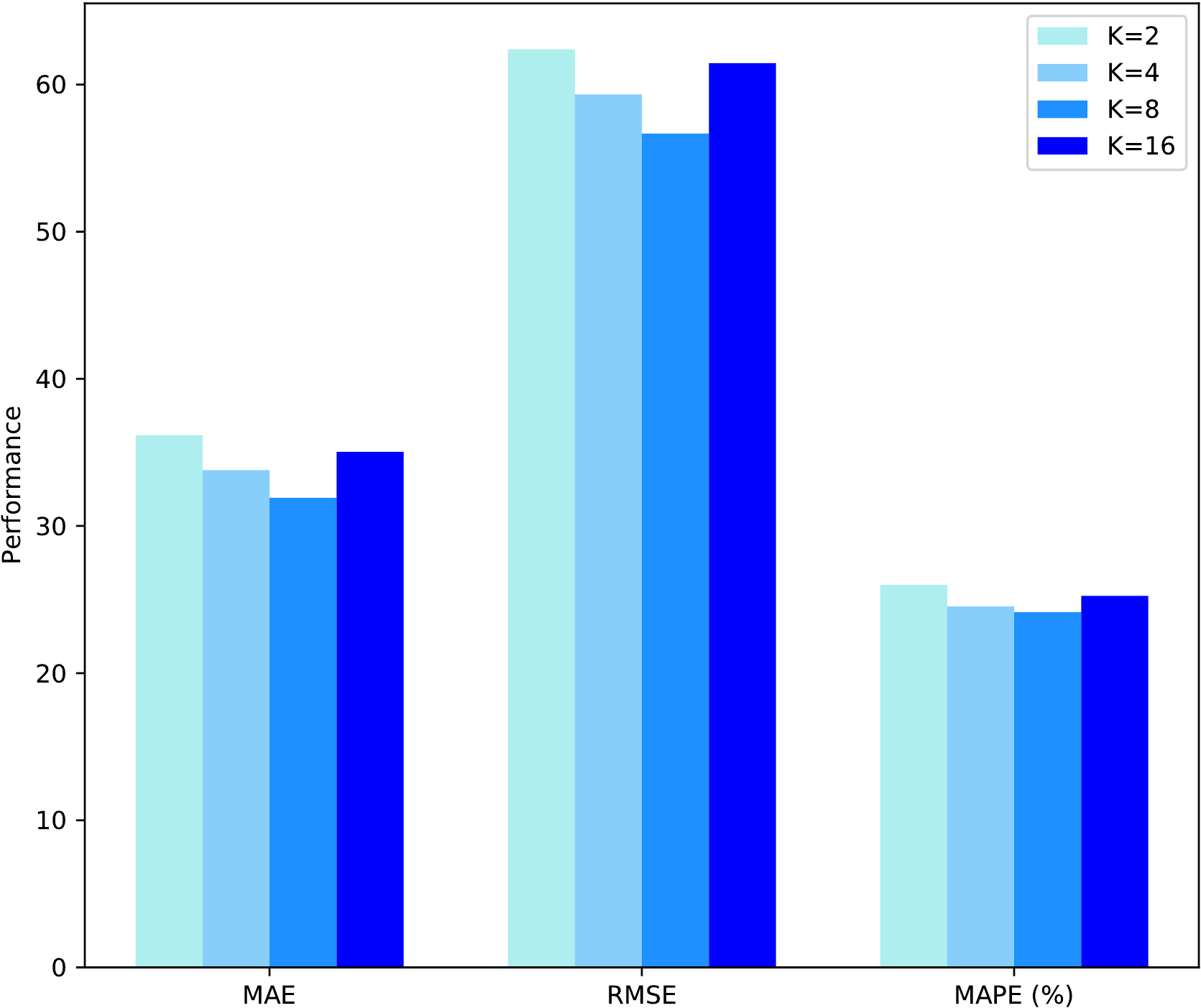}
    \caption{Linear attention heads.}
  \end{subfigure}%
  \hfill
  \begin{subfigure}{0.48\linewidth}
    \includegraphics[width=\linewidth]{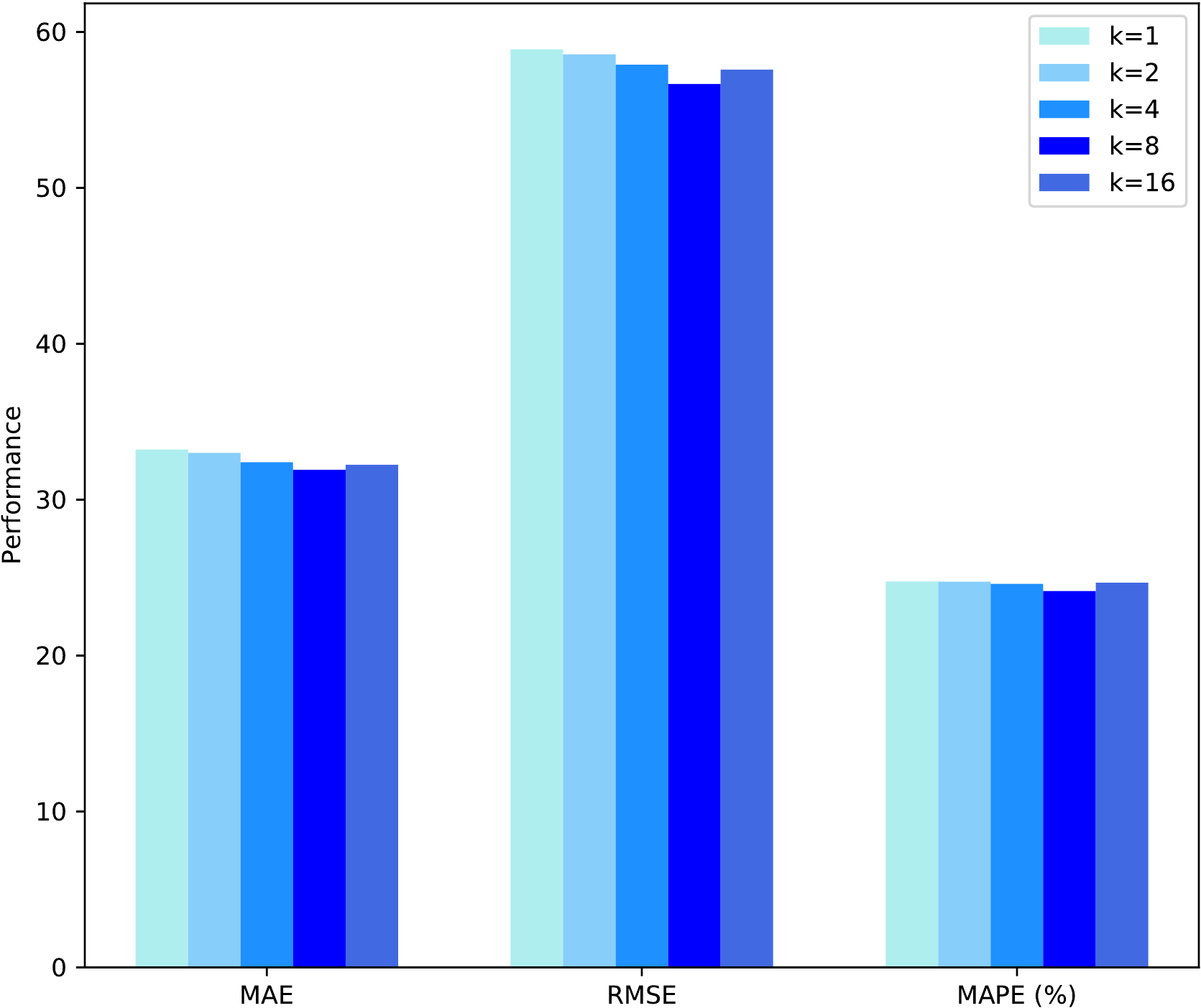}
    \caption{MHDCN heads.}
  \end{subfigure}%
  
  \begin{subfigure}{0.48\linewidth}
    \includegraphics[width=\linewidth]{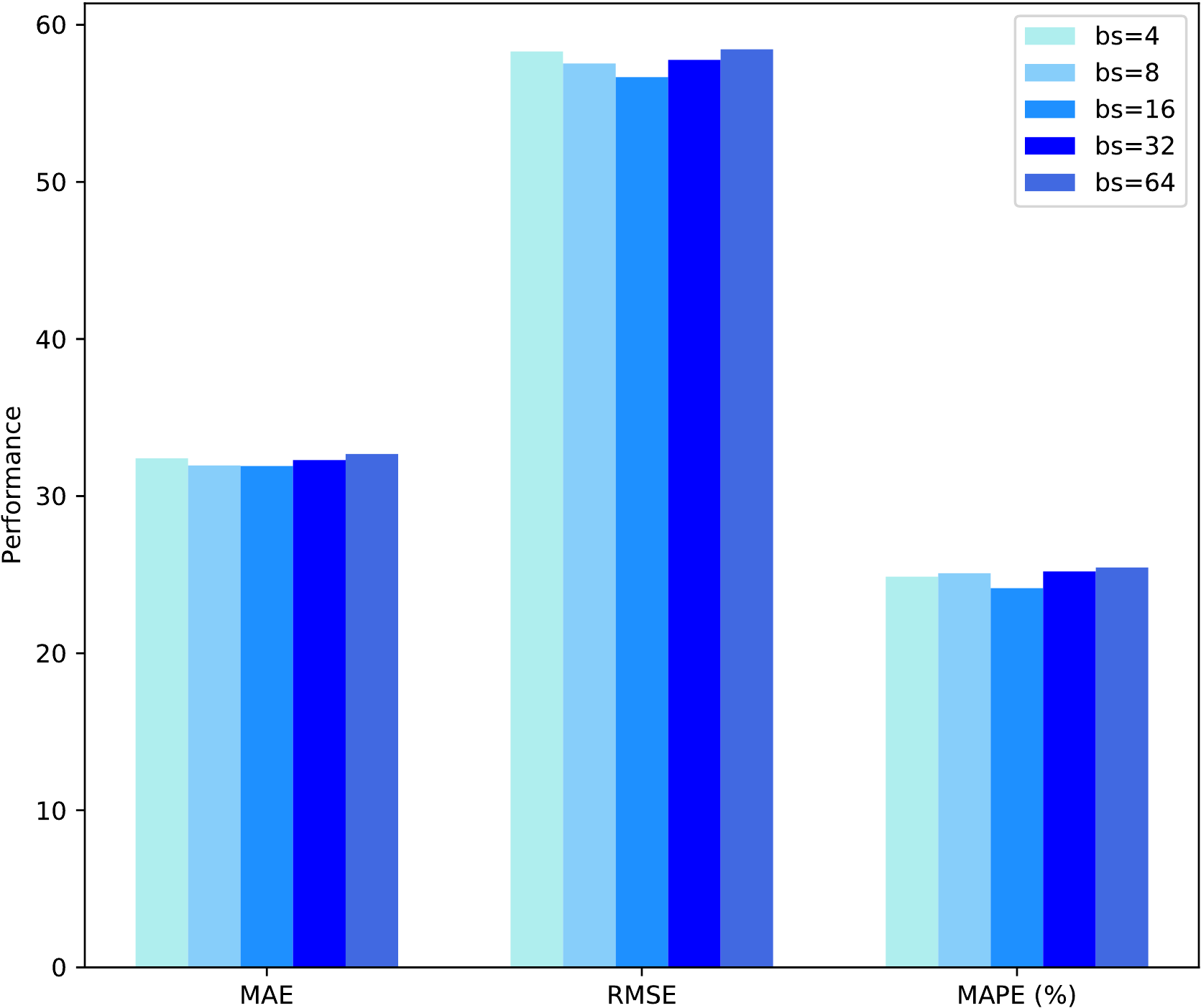}
    \caption{Batchsize.}
  \end{subfigure}%
  \hfill
  \begin{subfigure}{0.48\linewidth}
    \includegraphics[width=\linewidth]{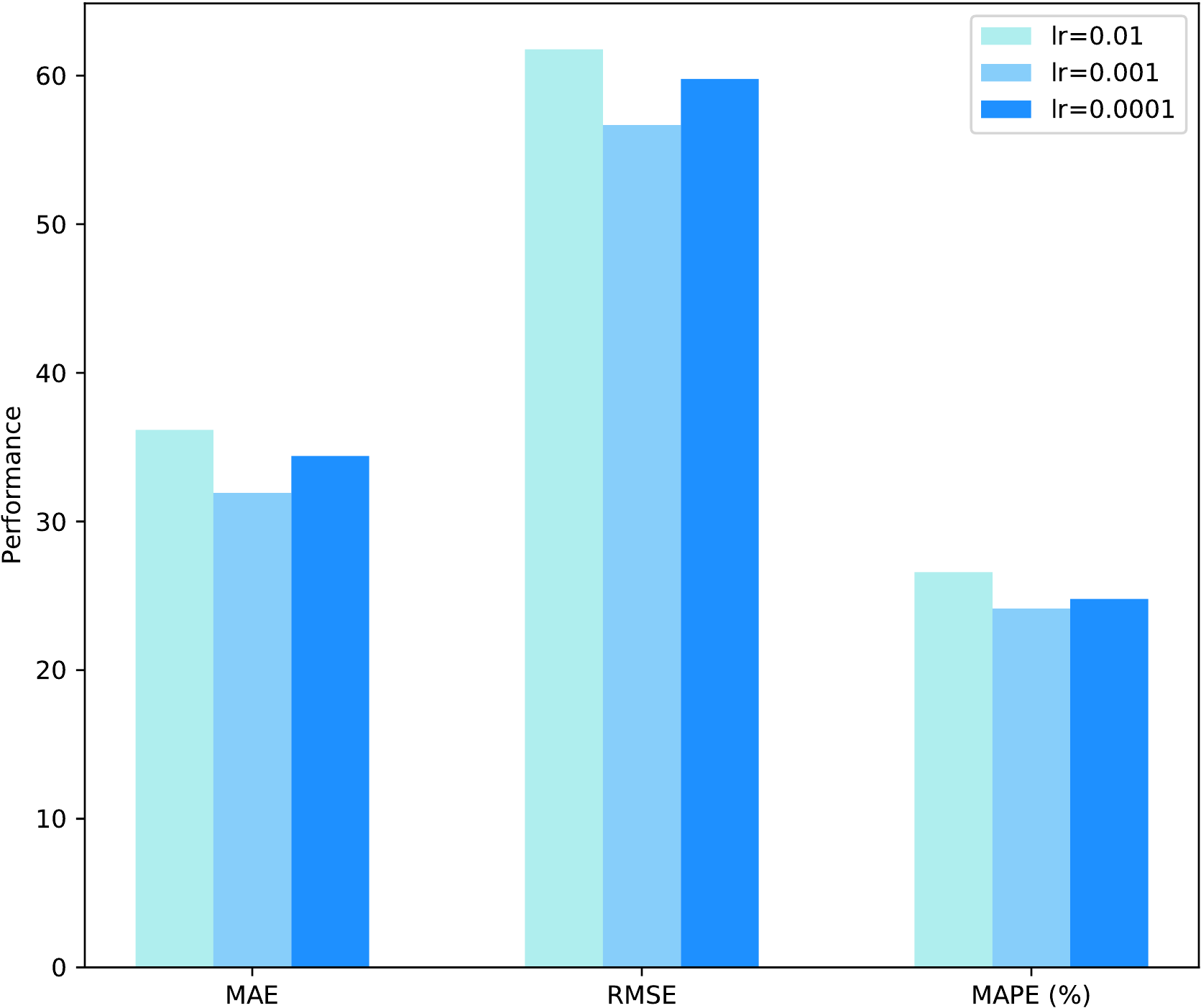}
    \caption{Learning rate.}
  \end{subfigure}%
  \caption{Experimental results under different hyperparameter settings.}
\end{figure}
\begin{table*}[t]
    \centering
    \caption{Performance comparison of MHDCN and other methods on the England dataset.}
    \begin{tabular}{cccc|ccc|ccc|ccc}
    \toprule
        \multirow{2}{*}{Model} & \multicolumn{3}{c|}{1 hour} & \multicolumn{3}{c|}{2 hour} & \multicolumn{3}{c|}{3 hour} & \multicolumn{3}{c}{Average}  \\
    \cmidrule{2-13}
         & MAE & RMSE & MAPE & MAE & RMSE & MAPE & MAE & RMSE & MAPE & MAE & RMSE & MAPE\\
    \midrule
        JKNet & 30.12 & 54.02 & 22.01\% & 35.01 & 62.17 & 28.30\% & 38.58 & 68.04 & 31.90\% & 32.51 & 58.28 & 24.76\%\\
        Mixhop & 30.06 & 54.01 & 22.03\% & 34.63 & 62.04 & 28.69\% & 37.42 & 66.63 & 32.06\% & 32.15 & 57.94 & 25.03\%\\
        MHGCN & 29.82 & 53.08 & 22.38\% & 34.28 & 60.89 & 27.81\% & 37.63 & 66.47 & 31.48\% & 32.01 & 57.17 & 24.71\%\\
        MHDCN (ours) & 29.79 & 53.32 & 20.74\% & 34.28 & 61.00 & 27.44\% & 37.49 & 66.11 & 31.00\% & 31.92 & 57.20 & 23.55\%\\
    \bottomrule
    \end{tabular}
\end{table*}
\begin{table*}[t]
    \centering
    \caption{Performance comparison of GRU and other methods on the England dataset.}
    \begin{tabular}{cccc|ccc|ccc|ccc}
    \toprule
        \multirow{2}{*}{Model} & \multicolumn{3}{c|}{1 hour} & \multicolumn{3}{c|}{2 hour} & \multicolumn{3}{c|}{3 hour} & \multicolumn{3}{c}{Average}  \\
    \cmidrule{2-13}
         & MAE & RMSE & MAPE & MAE & RMSE & MAPE & MAE & RMSE & MAPE & MAE & RMSE & MAPE\\
    \midrule
        RNN & 31.68 & 56.29 & 22.59\% & 36.09 & 64.34 & 28.42\% & 39.21 & 69.57 & 32.03\% & 33.74 & 60.35 & 25.21\%\\
        LSTM & 30.04 & 53.83 & 21.90\% & 34.46 & 61.51 & 27.94\% & 37.50 & 66.69 & 31.73\% & 32.12 & 57.72 & 24.66\%\\
        GRU (ours) & 29.79 & 53.32 & 20.74\% & 34.28 & 61.00 & 27.44\% & 37.49 & 66.11 & 31.00\% & 31.92 & 57.20 & 23.55\%\\
    \bottomrule
    \end{tabular}
\end{table*}
\subsection{Effect of Hyperparameters}
Fig. 4 shows the average prediction results of the model under different hyperparameter settings over next three hours on the England dataset. When we tune one of the hyperparameters, the other hyperparameters remain the default optimal values (K=8, k=8 d=16 and L=2). Fig. 4 (a), (b), (c), and (d) denote the network structure hyperparameters: the GRU layers, the dimensions of each head, the linear attention heads and the MHDCN heads. Fig. 4 (e) and (f) are model training hyperparameters: the batchsize and the learning rate. As shown in the results of the network structure hyperparameters, when the model is small, it is easy to underfit, and when the model becomes large, it is easy to overfit. Furthermore, we found the model is not sensitive to change the number of MHDCN heads, which proves that local information is to provide inductive bias to assist the model to train instead of the main part of the model. Besides, during model training, a larger batchsize speeds up model training, but the performance is worse. The batchsize with moderate speed and accuracy should be selected. Finally, we found that when the learning rate is too large; it is easy to skip the optimal solution, and when the learning rate is too small, it is easy to fall into the local optimal solution. We also should maintain a moderate learning rate and reduce it during training to achieve the best results.
\subsection{Dynamic Spatial Semantic Study}
Table \uppercase\expandafter{\romannumeral5} shows the performance when replacing the MHDCN in STJLA with other graph neural networks on England dataset. By comparing the results of MHGCN and MHDCN, we can see that the road network cannot be regarded as an undirected graph because the traffic flow is irreversible. At the same time, we implemented a three layers Mixhop \cite{abu2019mixhop} with the Delta Operator of 0.05 and a three layers JKNet with the Max-pooling. Mixhop achieved better performance, proving that increasing the graph convolution width is better than increasing the depth to increase receptive field. Therefore, we propose the MHDCN that does not slow down model training while increasing the model width.
\subsection{Dynamic Temporal Semantic Study}
Table \uppercase\expandafter{\romannumeral6} shows the performance when replacing the GRU in STJLA with other recurrent neural networks on England dataset. The RNN has no gating mechanism and can neither capture long-term temporal dependence nor solve the problems of gradient vanish, and achieves the worst performance. Both GRU and LSTM have designed good gating mechanisms to capture long-term temporal dependence, and they have achieved better results than RNN. We choose GRU to learn dynamic temporal semantic context because it is simpler and more efficient than LSTM.
\subsection{Case Study}
In this section, we conduct a case study to intuitively show the performance of STJLA. We select road 20 and road 104 in the England dataset and plot the traffic flow prediction results of last week based on VAR, DCRNN and STJLA in Fig. 5. The results in Fig. 5 (a) and (b) prove that traditional statistical methods can only model linear dependence and cannot make good predictions for nonlinear data. Compared with VAR, the deep learning method DCRNN has a significant increase to extract nonlinear temporal dependence, i.e., prediction value is no longer just a straight line but also a curve. However, DCRNN has insufficient ability to predict fluctuations of different region nodes. We observed STJLA captures the temporal trend of traffic flow more accurately than the other two methods of road 20 and road 104. It shows that STJLA could better capture the structure and semantic correlations of the road network and time to make a more accurate prediction.
\begin{figure}[htbp]
\centering
    \begin{subfigure}{0.48\linewidth}
    \includegraphics[width=\linewidth]{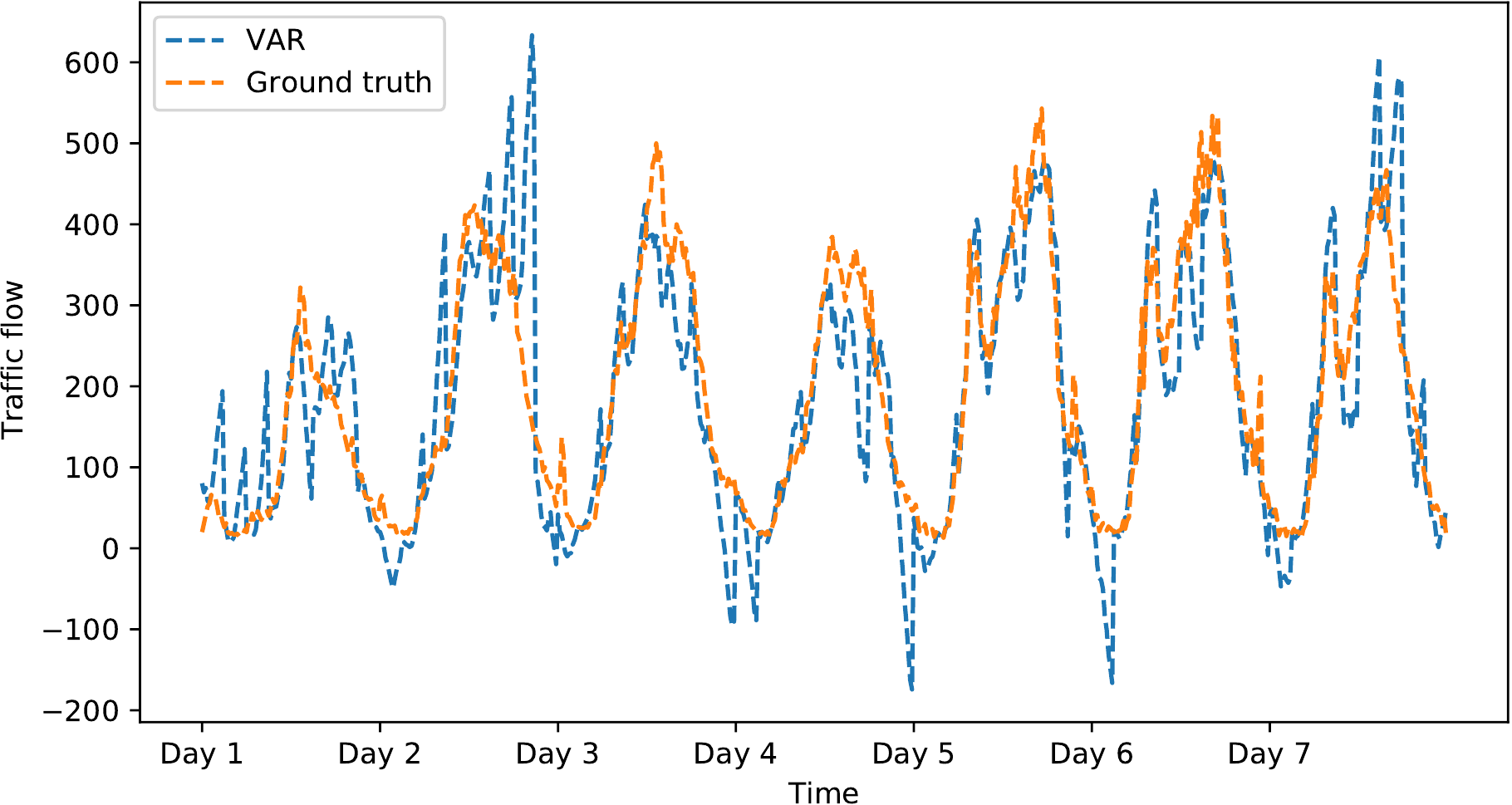}
    \caption{Road 20.}
  \end{subfigure}%
  \hfill
  \begin{subfigure}{0.48\linewidth}
    \includegraphics[width=\linewidth]{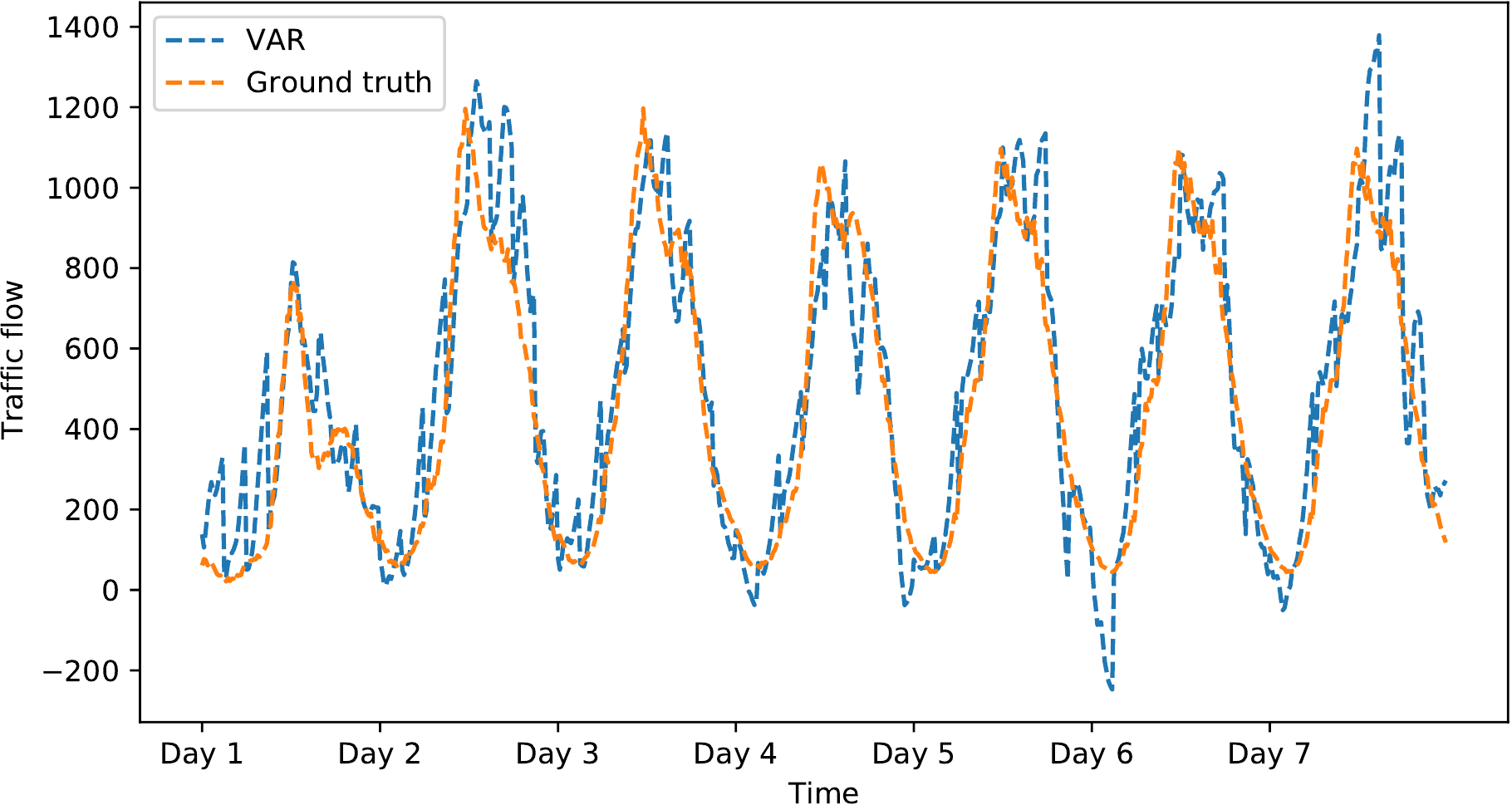}
    \caption{Road 104.}
  \end{subfigure}
  
  \begin{subfigure}{0.48\linewidth}
    \includegraphics[width=\linewidth]{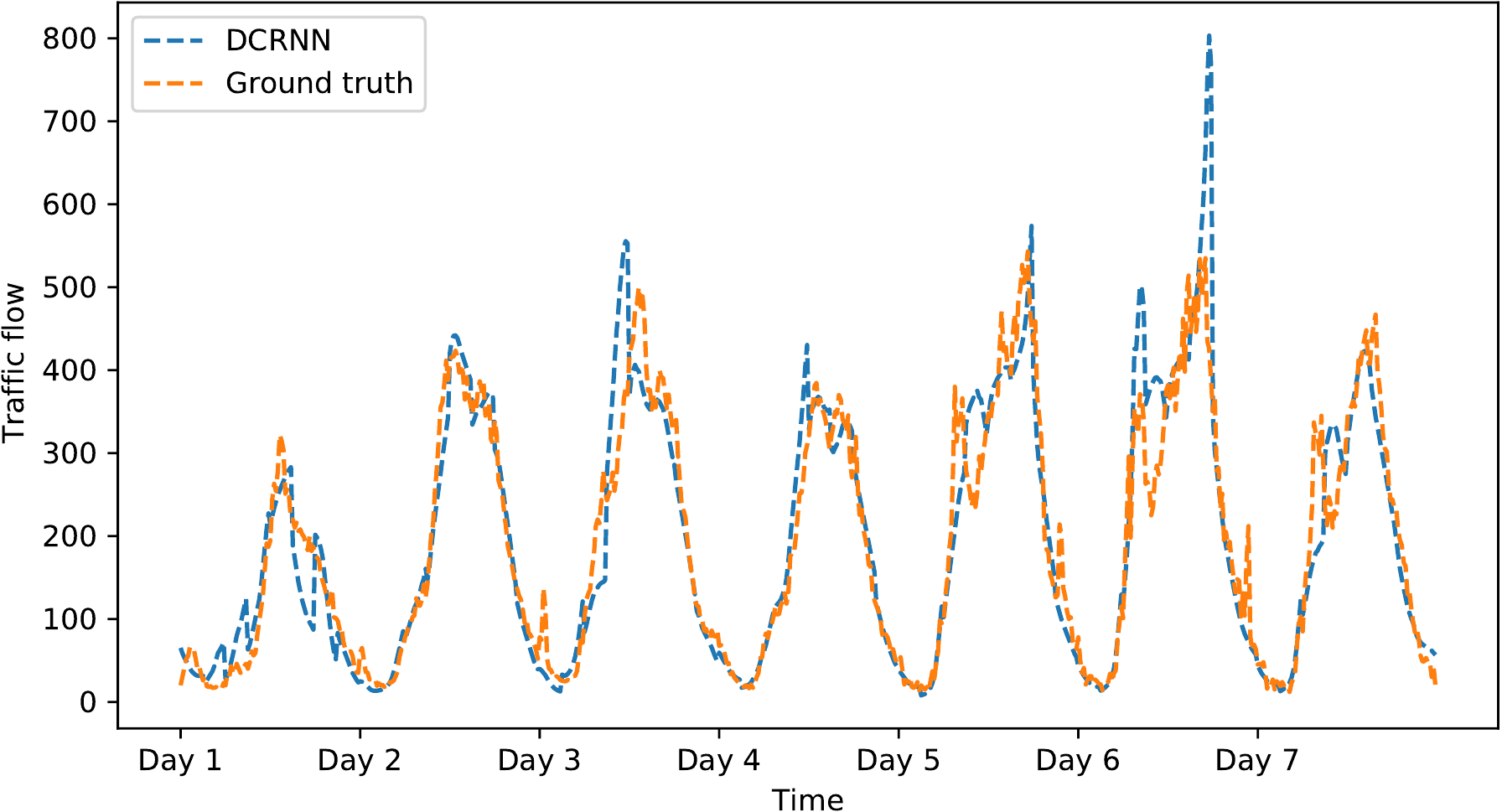}
    \caption{Road 20.}
  \end{subfigure}%
  \hfill
  \begin{subfigure}{0.48\linewidth}
    \includegraphics[width=\linewidth]{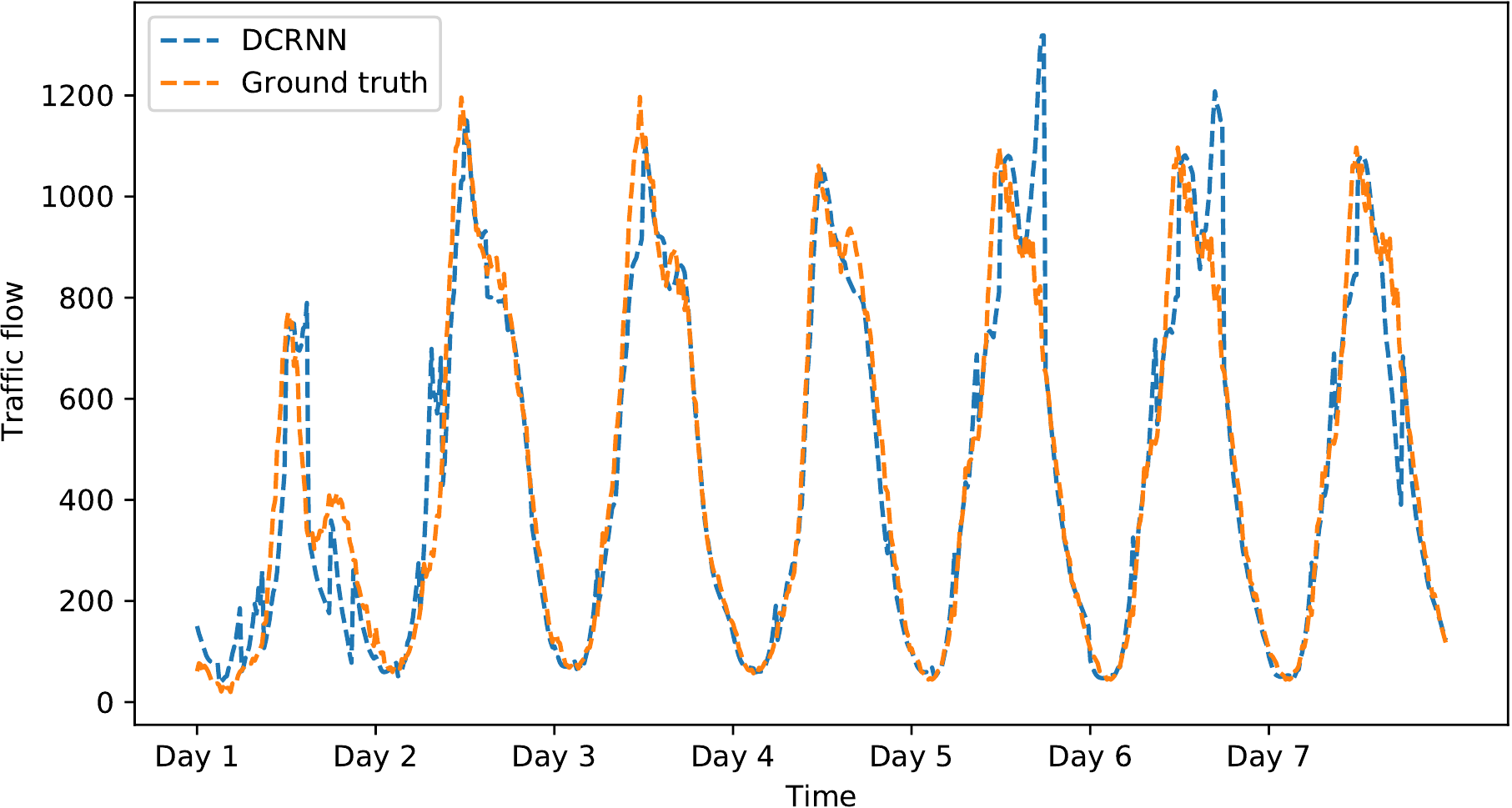}
    \caption{Road 104.}
  \end{subfigure}
  
  \begin{subfigure}{0.48\linewidth}
    \includegraphics[width=\linewidth]{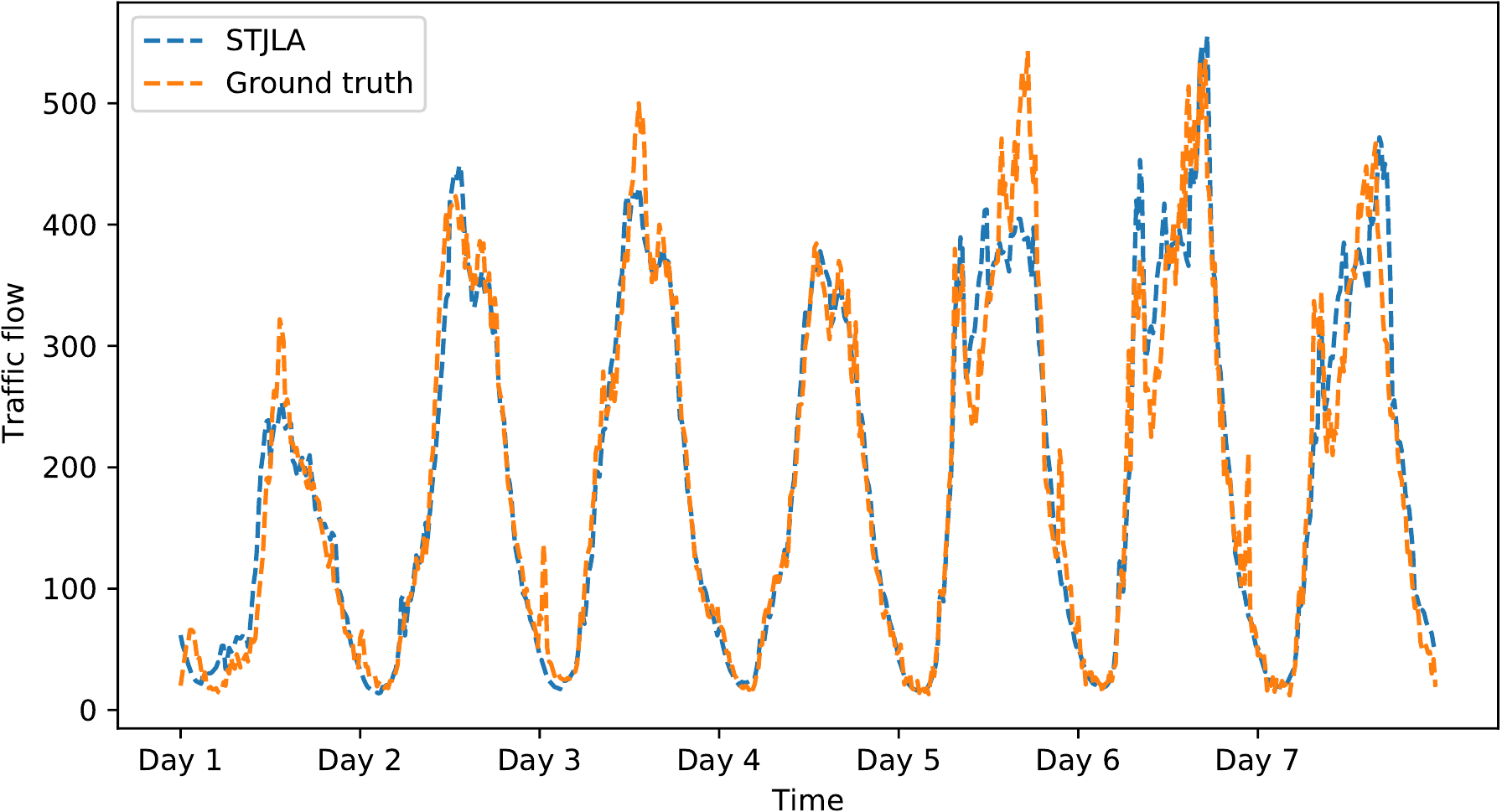}
    \caption{Road 20.}
  \end{subfigure}%
  \hfill
  \begin{subfigure}{0.48\linewidth}
    \includegraphics[width=\linewidth]{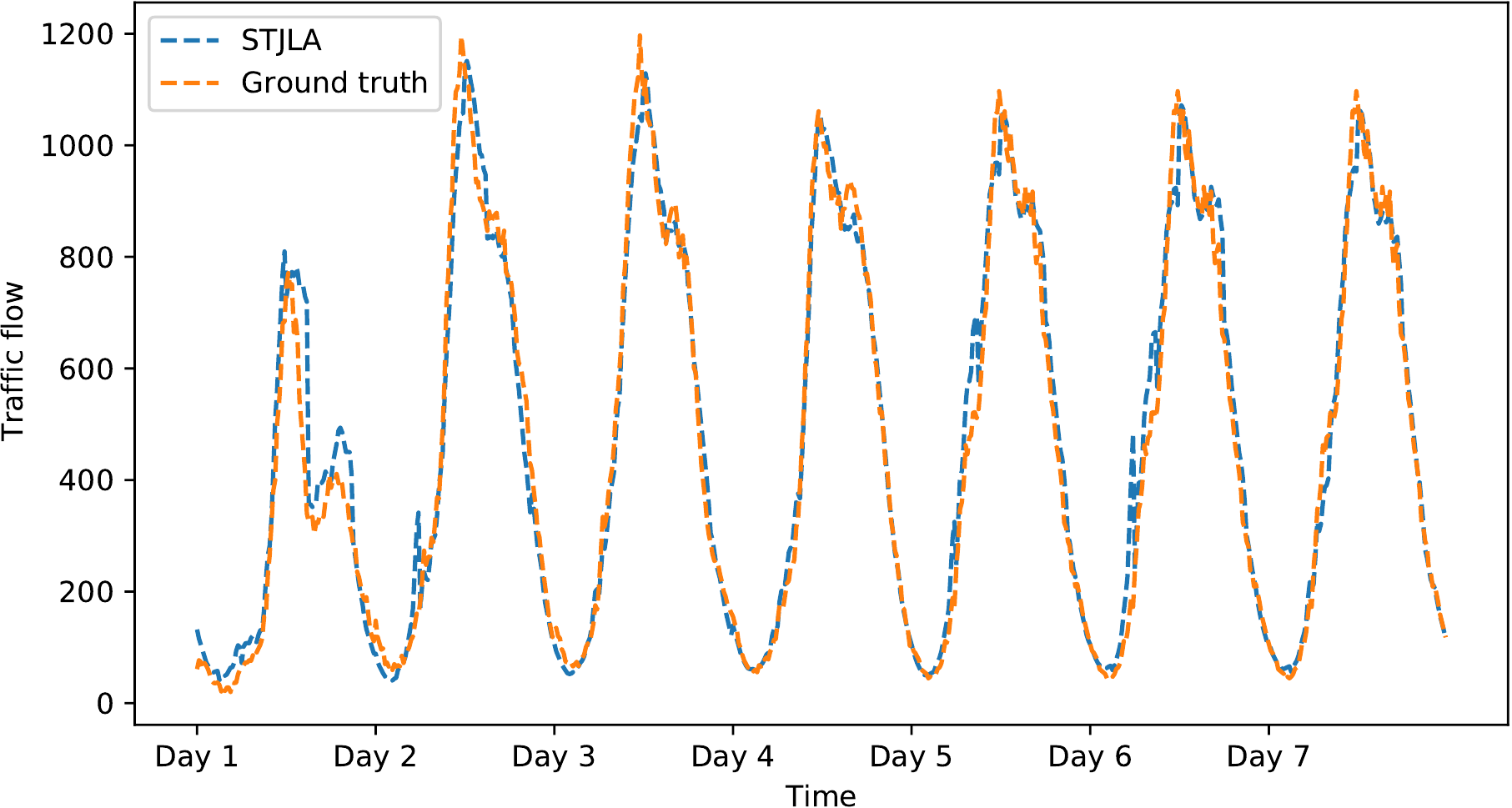}
    \caption{Road 104.}
  \end{subfigure}
  \caption{A case study of traffic flow prediction.}
\end{figure}
\subsection{Complexity Analysis}
In this section, we report on the time consumption and memory usage of STJLA and baselines with the England dataset, as shown in Table \uppercase\expandafter{\romannumeral7}. In terms of the training time, we find our STJLA is faster than RNN-based models and attention-based models, such as DCRNN and GMAN, because RNN-based models cannot be parallelized and attention-based models have $O(TN^2+NT^2)$ time complexity. GWN and MTGNN perform best in both training and testing stages because they are not auto-regressive models. Besides, models such as STGCN, DCRNN and ST-GRAT inference by dynamic decoding have slow inference speed. We use linear attention and generative style inference to speed up STJLA training and inference speed and reduce memory consumption.
\begin{table}[htbp]
    \centering
    \caption{The computation times and memory on the England dataset.}
    \resizebox{1.0\linewidth}{!}{\begin{tabular}{ccccccccc}
    \toprule
        & DCRNN & STGCN & GWN & AGCRN & MTGNN & GMAN & ST-GRAT & STJLA\\
    \midrule
        Training time (s/epoch) & 1029 & 226 & 213 & 1847 & 200 & 534 & 737 & 269\\
        Inference time (s) & 132 & 223 & 30 & 458 & 17 & 53 & 277 & 38\\
        Memory usage (MB) & 2843 & 1305 & 1615 & 3303 & 1361 & 10729 & 13643 & 4225\\
    \bottomrule
    \end{tabular}}
\end{table}
\section{Conclusion}
In this paper, we formulated the traffic prediction on road networks as a spatio-temporal forecasting problem, and proposed a multi-context aware spatio-temporal joint linear attention network. The novel linear attention STJLA designed captures the spatio-temporal dependencies between all the sensors in the joint spatio-temporal graph. Specifically, we use node2vec and one-hot encoding algorithm to extract static context from structure and we further utilize MHDCN and GRU to mine dynamic context from semantics. The multi-scale context can improve performance of spatio-temporal joint linear attention. We further design a transform layer to convert historical states into future states. Evaluation on two large-scale real-world traffic datasets demonstrates that our proposed STJLA achieves state-of-the-art results. In future work, we will apply the STJLA to other spatio-temporal prediction tasks, such as video prediction.

%



\section*{Acknowledgment}
Many thanks to the Beijing Key Laboratory of Mobile Computing and Pervasive Device, Institute of Computing Technology Chinese Academy of Sciences for the support in our research.




\bibliographystyle{IEEEtran}
\bibliography{HEFGAT}
%



%
\vspace{-35pt}
\begin{IEEEbiography}[{\includegraphics[width=1in,height=1.25in,clip,keepaspectratio]{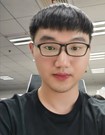}}]{Yuchen Fang}
received the B.S. degree from the School of Information Science and Technology, Beijing Forestry University, Beijing, China. He is currently working toward the M.S. degree with the School of Computer Science (National Pilot Software Engineering School), Beijing University of Posts and Telecommunications, Beijing, China. His current main interests include traffic forecasting and graph neural network.
\end{IEEEbiography}
\vspace{-45pt}
\begin{IEEEbiography}[{\includegraphics[width=1in,height=1.25in,clip,keepaspectratio]{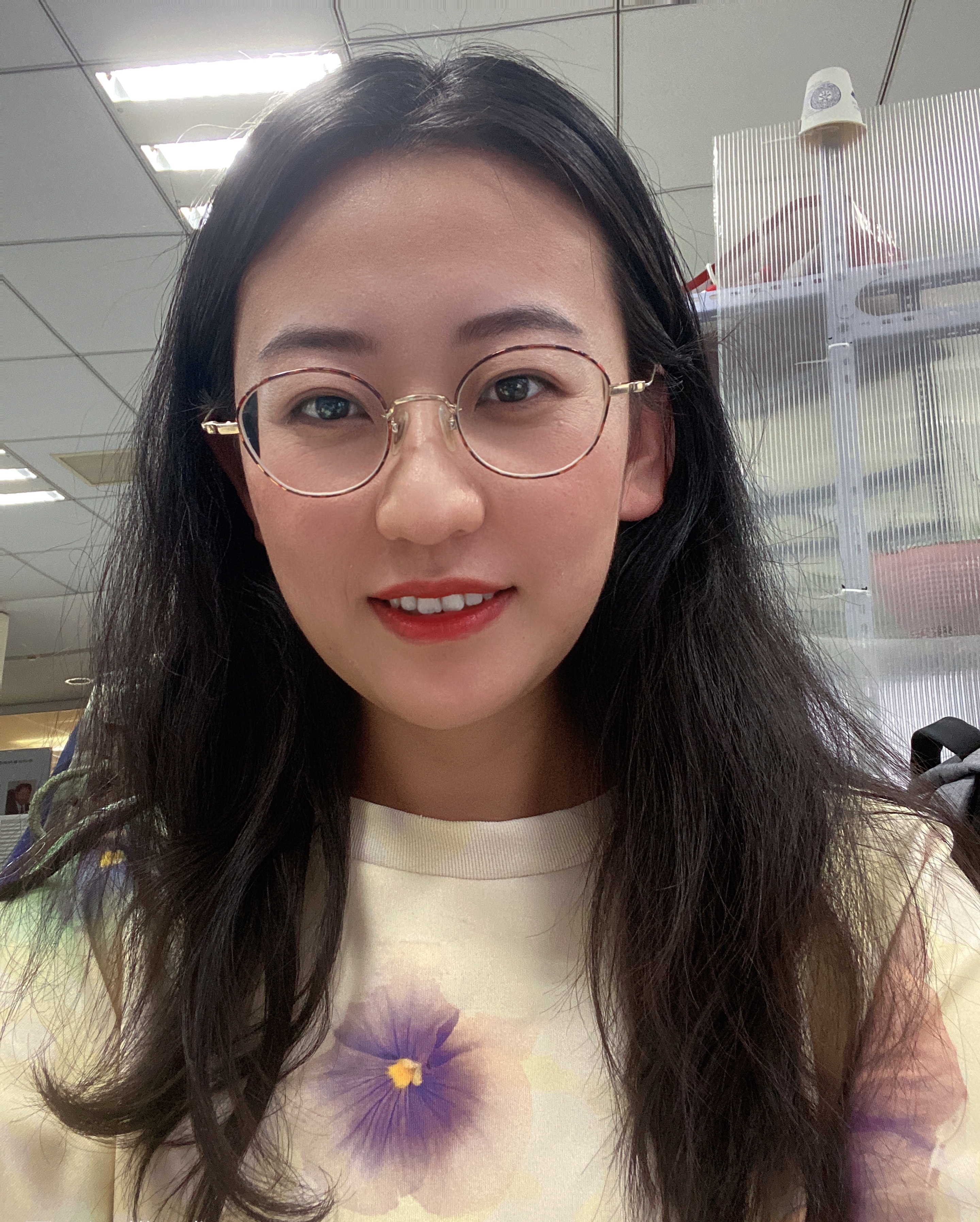}}]{Yanjun Qin}
is currently working toward the Ph.D. degree with the School of Computer Science (National Pilot Software Engineering School), Beijing University of Posts and Telecommunications, Beijing, China. Her current main interests include location based services, pervasive computing, convolution neural networks, and machine learning. He is mainly involved in traffic pattern recognition related project research and implementation.
\end{IEEEbiography}
\vspace{-45pt}
\begin{IEEEbiography}[{\includegraphics[width=1in,height=1.25in,clip,keepaspectratio]{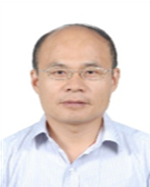}}]{Haiyong Luo}
received the B.S. degree from the Department of Electronics and Information Engineering, Huazhong University of Science and Technology, Wuhan, China, in 1989, the M.S. degree from the School of Information and Communication Engineering, the Beijing University of Posts and Telecommunication China, Beijing, China, in 2002, and Ph.D. degree in computer science from the University of Chines Academy of Sciences, Beijing, China, in 2008. He is currently an Associate Professor with the Institute of Computer Technology, Chinese Academy of Science, Beijing, China. His main research interests are location based services, pervasive computing, mobile computing, and Internet of Things.
\end{IEEEbiography}
\vspace{-45pt}
\begin{IEEEbiography}[{\includegraphics[width=1in,height=1.25in,clip,keepaspectratio]{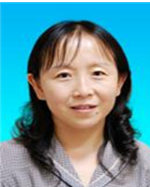}}]{Fang Zhao} received the B.S. degree from the School of Computer Science and Technology, Huazhong University of Science and Technology, Wuan, China, in 1990, and the M.S. and Ph.D. degrees in computer science and technology from the Beijing University of Posts and Telecommunication, Beijing, China, in 2004 and 2009, respectively. She is currently a Professor with the School of Computer Science (National Pilot Software Engineering School), Beijing University of Posts and Telecommunications, Beijing, China. Her research interests include mobile computing, location based services, and computer networks.
\end{IEEEbiography}
\vspace{-45pt}
\begin{IEEEbiography}[{\includegraphics[width=1in,height=1.25in,clip,keepaspectratio]{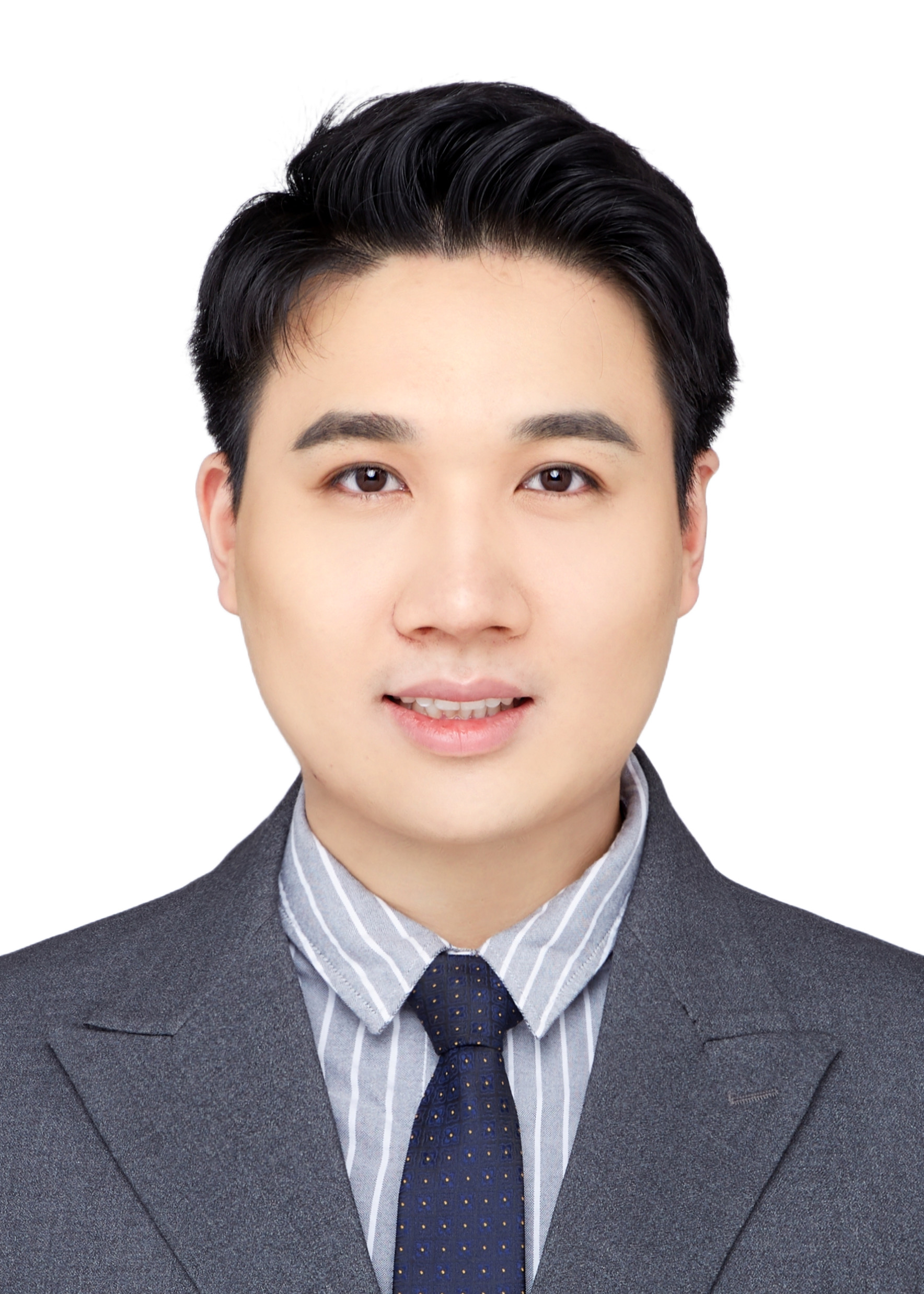}}]{Chenxing Wang}
is currently pursuing the P.h.D with the School of Computer Science (National Pilot Software Engineering School), Beijing University of Posts and Telecommunications, Beijing, China. His current main interests include spatial-temporal data mining, travel time estimation, traffic flow prediction and transportation mode detection using deep learning techniques.
\end{IEEEbiography}
\end{document}